\documentclass[10pt,a4paper]{article}
\usepackage[T1]{fontenc}
\usepackage[utf8]{inputenc}
\usepackage{newtxtext,newtxmath}
\usepackage[scaled=0.94]{helvet}
\usepackage[a4paper,textwidth=176mm,top=18mm,bottom=19mm,headheight=13pt,headsep=7mm,footskip=9mm]{geometry}
\usepackage{microtype}
\usepackage{graphicx,booktabs,array,longtable,ragged2e}
\usepackage[table]{xcolor}
\usepackage{caption,titlesec,fancyhdr,placeins,enumitem}
\usepackage[numbers,sort&compress]{natbib}
\usepackage[hyphens]{url}
\usepackage[unicode,colorlinks=true,linkcolor=linkink,citecolor=linkink,urlcolor=linkink,bookmarksnumbered=true]{hyperref}
\usepackage{bookmark}
\definecolor{linkink}{HTML}{165A67}
\definecolor{tabshade}{HTML}{F3F6F6}
\titleformat{\section}{\sffamily\bfseries\large}{\thesection}{0.7em}{}
\titleformat{\subsection}{\sffamily\bfseries\normalsize}{\thesubsection}{0.7em}{}
\titlespacing*{\section}{0pt}{15pt plus 2pt minus 2pt}{6pt}
\titlespacing*{\subsection}{0pt}{10pt plus 2pt minus 1pt}{4pt}
\fancypagestyle{plain}{\fancyhf{}\fancyfoot[C]{\sffamily\small\thepage}}
\renewenvironment{abstract}{\par\begin{center}\sffamily\bfseries Abstract\end{center}\vspace{-5pt}\begingroup\fontsize{10.5}{12.6}\selectfont\setlength{\parskip}{1pt}}{\par\endgroup}
\hypersetup{pdftitle={Autonomous Structuring of Radiology Reports Across Modalities at Archive Scale Using an Open-Weight Large Language Model},pdfauthor={Friedrich Puttkammer et al.},pdfsubject={Unblinded preprint with supplementary material},pdfkeywords={radiology, structured reporting, large language models, RadFields}}
\begin{document}
\fontsize{10.5}{13}\selectfont

\thispagestyle{plain}
\begin{center}
{\sffamily\bfseries\fontsize{20}{23}\selectfont Autonomous Structuring of Radiology Reports Across Modalities at Archive Scale Using an Open-Weight Large Language Model\par}
\vspace{11pt}
{\fontsize{10.5}{13.5}\selectfont Friedrich Puttkammer\textsuperscript{1,2}, Fabian Drexel\textsuperscript{2,3}, Marlene Fritzsche\textsuperscript{2}, Era Stambollxhiu\textsuperscript{2}, Miriam Kumpf\textsuperscript{2}, Lena Schmitzer\textsuperscript{2}, Lea Schumann\textsuperscript{2}, Lina Xu\textsuperscript{1}, Johannes Moll\textsuperscript{2,3}, Jannik Lübberstedt\textsuperscript{2}, Zeineb Ben Chaaben\textsuperscript{2}, Anirudh Narayanan\textsuperscript{1,2}, Hartmut Häntze\textsuperscript{1,2}, Renato Cuocolo\textsuperscript{4}, Antonios Billis\textsuperscript{5}, Alexander Löser\textsuperscript{6}, Jawed Nawabi\textsuperscript{7,8}, Marcus R. Makowski\textsuperscript{2}, Cosmin I. Bercea, Shahrooz Faghihroohi\textsuperscript{9}, Lisa C. Adams\textsuperscript{2*}, Keno K. Bressem\textsuperscript{2,10,11,12*}\par}
\vspace{9pt}
\begin{minipage}{\textwidth}\fontsize{8.5}{10.3}\selectfont\RaggedRight\hyphenpenalty=10000\exhyphenpenalty=10000
\hangindent=1.3em\hangafter=1\noindent\makebox[1.3em][l]{\textsuperscript{1}}Department of Radiology, Charité -- Berlin University of Medicine, corporate member of Freie Universität Berlin and Humboldt Universität zu Berlin, Berlin, Germany\par\vspace{1pt}
\hangindent=1.3em\hangafter=1\noindent\makebox[1.3em][l]{\textsuperscript{2}}Department of Diagnostic and Interventional Radiology, School of Medicine and Health, TUM Klinikum Rechts der Isar, Technical University of Munich, Munich, Germany\par\vspace{1pt}
\hangindent=1.3em\hangafter=1\noindent\makebox[1.3em][l]{\textsuperscript{3}}Chair for AI in Healthcare and Medicine, Technical University of Munich (TUM) and TUM University Hospital, Munich, Germany\par\vspace{1pt}
\hangindent=1.3em\hangafter=1\noindent\makebox[1.3em][l]{\textsuperscript{4}}Department of Medicine, Surgery, and Dentistry, University of Salerno, Salerno, Italy\par\vspace{1pt}
\hangindent=1.3em\hangafter=1\noindent\makebox[1.3em][l]{\textsuperscript{5}}Lab of Medical Physics \& Digital Innovation, School of Medicine, Aristotle University of Thessaloniki, Thessaloniki, Greece\par\vspace{1pt}
\hangindent=1.3em\hangafter=1\noindent\makebox[1.3em][l]{\textsuperscript{6}}Department of Computer Science and Media, Berlin University of Applied Sciences and Technology, Berlin, Germany\par\vspace{1pt}
\hangindent=1.3em\hangafter=1\noindent\makebox[1.3em][l]{\textsuperscript{7}}Department of Neuroradiology, Charité -- Berlin University of Medicine, corporate member of Freie Universität Berlin and Humboldt Universität zu Berlin, Berlin, Germany\par\vspace{1pt}
\hangindent=1.3em\hangafter=1\noindent\makebox[1.3em][l]{\textsuperscript{8}}Berlin Institute of Health (BIH), BIH Biomedical Innovation Academy, Berlin, Germany\par\vspace{1pt}
\hangindent=1.3em\hangafter=1\noindent\makebox[1.3em][l]{\textsuperscript{9}}Chair for Computer Aided Medical Procedures \& Augmented Reality, Technical University of Munich (TUM) and TUM University Hospital, Munich, Germany\par\vspace{1pt}
\hangindent=1.3em\hangafter=1\noindent\makebox[1.3em][l]{\textsuperscript{10}}Institute for Artificial Intelligence in Medicine (IKIM), University Hospital Essen (AöR), Essen, Germany\par\vspace{1pt}
\hangindent=1.3em\hangafter=1\noindent\makebox[1.3em][l]{\textsuperscript{11}}Institute of Interventional and Diagnostic Radiology and Neuroradiology, University Hospital Essen, Essen, Germany\par\vspace{1pt}
\hangindent=1.3em\hangafter=1\noindent\makebox[1.3em][l]{\textsuperscript{12}}National Center for Tumor Diseases (NCT), NCT West, a partnership between DKFZ and University Hospital Essen, Essen, Germany\par\vspace{1pt}
\end{minipage}
\vspace{6pt}
{\fontsize{8.2}{10}\selectfont\textsuperscript{*}These authors contributed equally as senior authors.\quad Correspondence: \href{mailto:keno.bressem@nct-west.de}{keno.bressem@nct-west.de}\par}
\end{center}
\vspace{1pt}
\begin{abstract}
\noindent\textbf{Purpose:} To develop and evaluate an open-weight large language model (LLM) pipeline that converts an entire archive of free-text radiology reports into structured reports without human oversight.\par\smallskip
\noindent\textbf{Materials and Methods:} In this retrospective study, a pipeline with 150 hierarchically organized templates was developed at one center and tested at a second center on reports from 2010 to 2025. The open-weight model gpt-oss-120B selects the template in three constrained-decoding steps and fills it on one local graphics processing unit. Template selection was scored against expert labels on 914 randomly sampled reports of five modalities, structuring quality on 920 radiography and CT reports corrected field by field by five residents. The pipeline then processed the complete archive of the second center. Proportions are reported with Wilson 95\% confidence intervals (CIs).\par\smallskip
\noindent\textbf{Results:} An optimal template set was selected for 74.4\% of reports (680 of 914; 95\% CI: 71.5\%, 77.1\%) and an appropriate set for 82.3\% (752 of 914; 95\% CI: 79.7\%, 84.6\%), 87.7\% for single-region and 54.1\% for multi-region reports. Macro semantic textual similarity between output and corrected reference was 0.95 for radiography and 0.97 for CT, residents left 88.7\% of 24,638 fields unchanged, and unsupported content was flagged in 1.0\% and 1.5\% of reports. The pipeline produced structured output for 96.5\% of 2,186,982 archive reports (2,401,544 structured reports) at 1,258 reports per hour on one graphics processing unit.\par\smallskip
\noindent\textbf{Conclusion:} An open-weight LLM pipeline structured a complete multimodality report archive without human oversight with high content fidelity. Multi-region reports remained the main source of template errors.\par\smallskip
\end{abstract}
\clearpage

\section{Introduction}

A radiology archive holds millions of free-text reports, representing a rich source of information for research and quality control. However, due to the free text nature, they remain inaccessible for automated evaluation \cite{ref1}. For this reason radiological societies have recommended structured reporting for more than a decade \cite{ref2}, yet most reports are still narratives whose format and terminology vary between institutions and radiologists \cite{ref3,ref4}. Structured reporting at the time of dictation would be the optimal solution, but longer reporting times, disagreement about templates, and disrupted dictation workflows slow down adoption \cite{ref5}. Structuring reports afterwards leaves the reporting workflow untouched and makes archived reports usable.

Given that large language models (LLMs) can process free text narratives, multiple studies investigated them to extract predefined variables from radiology reports \cite{ref6,ref7,ref8,ref9,ref10} and map report content onto structured templates \cite{ref11,ref12,ref13,ref14,ref15}. In nearly all of these studies the template was known in advance, because only one or a few examination types were studied. However, an archive contains hundreds of examination types, and metadata do not reliably tell which template a report needs. To scale to a whole archive, the structuring pipeline must act fully autonomously.

A proof-of-concept study with GPT-4 assigned 170 fictitious CT and MRI reports to one of a few templates and structured them \cite{ref16}, showing that an LLM can perform both steps without human oversight. Yet, transferring this to a real archive raises four problems. First, a template collection must cover most examinations a department performs, with shared findings represented identically so that analyses can aggregate across templates. Second, presenting an LLM with more than 100 templates at once is inefficient, and a flat multiple-choice design handles neither reports describing several examinations nor rare examinations or reports without findings. Third, examination types differ between institutions, so templates must be easy to add, change, and remove. Fourth, a whole archive is a large volume of sensitive data. Sending it to a commercial API exposes the archive to a third party and is costly.

The purpose of this study was to develop a pipeline that addresses these four problems. The pipeline runs locally with open-weight LLMs and without human oversight, assigns each report to one or more of 150 hierarchically organized templates or rejects it, and fills the selected template. We tested it in a reader study at a second center for template selection and structuring quality and applied it to that center's complete archive of about 2.1 million reports across five modalities.

\section{Materials and Methods}

\subsection{Study Design}

This retrospective study was approved by the institutional review boards of Charité -- Berlin University of Medicine (Charité; EA2/148/23) and Technical University of Munich (TUM; 2024-590-S-CB), which waived informed consent. Charité reports (January 1, 2009, to April 1, 2024) were used for template development and prompt engineering, TUM reports (January 1, 2010, to December 31, 2025) for the two test sets and for structuring the complete archive. The reports had not been used in prior studies, and patient demographics were not available at report level. Reports, templates and prompts were in German. The study follows the CLAIM 2024 checklist (Supplement 1) \cite{ref17}.

\begin{figure}[!htbp]
\centering
\includegraphics[width=\linewidth]{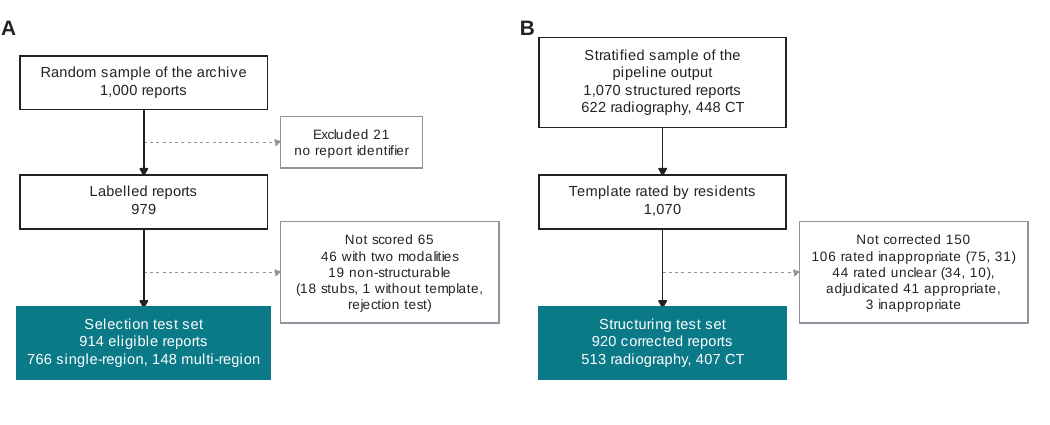}
\caption{Creation of the two test sets from the TUM archive. A, selection test set. Of a random sample of the archive, reports without a report identifier were excluded before labeling, reports combining two modalities and non-structurable reports (report stubs, examination without a template) were labeled but not scored, the non-structurable reports served as the rejection test. B, structuring test set. Of a stratified sample of the pipeline output with 9 to 25 reports per template, the residents rated the selected template, reports rated inappropriate or unclear were not corrected, the unclear ratings were adjudicated by two board-certified radiologists in consensus for the template rating. Numbers in parentheses are radiography, CT.}\label{fig:1}
\end{figure}

\subsection{Template Database}

To identify the examinations and diagnoses in the archive, an LLM (gpt-oss-120B) screened 10,000 random Charité reports for modality, body region and main diagnosis (Supplemental Methods S1), and two board-certified radiologists (KKB and LCA; 10 and 11 years of experience) defined 150 templates covering all common examination types from the resulting frequency tables. Each template defines the sections of a structured report and describes the expected content of each section.

Templates form a two-level hierarchy. A base template defines the sections of an examination type, for example CT chest, and derived templates add or replace sections for a clinical question. At run time the pipeline merges the derived template into its base template section by section, so structured reports of one examination type remain comparable across diagnoses (\hyperref[fig:S3]{Figure~S3}). The 150 templates (24 base, 75 derived, 51 independent without a base) cover five modalities and 37 examination types (\hyperref[tab:1]{Table~1}, \hyperref[fig:S1]{Figure~S1}). For CT reports that describe several examination types, 31 combination types define which examination types may occur together, and the report receives one template per examination type (Supplemental Methods S2).

\subsection{Pipeline}

\begin{figure}[!htbp]
\centering
\includegraphics[width=\linewidth]{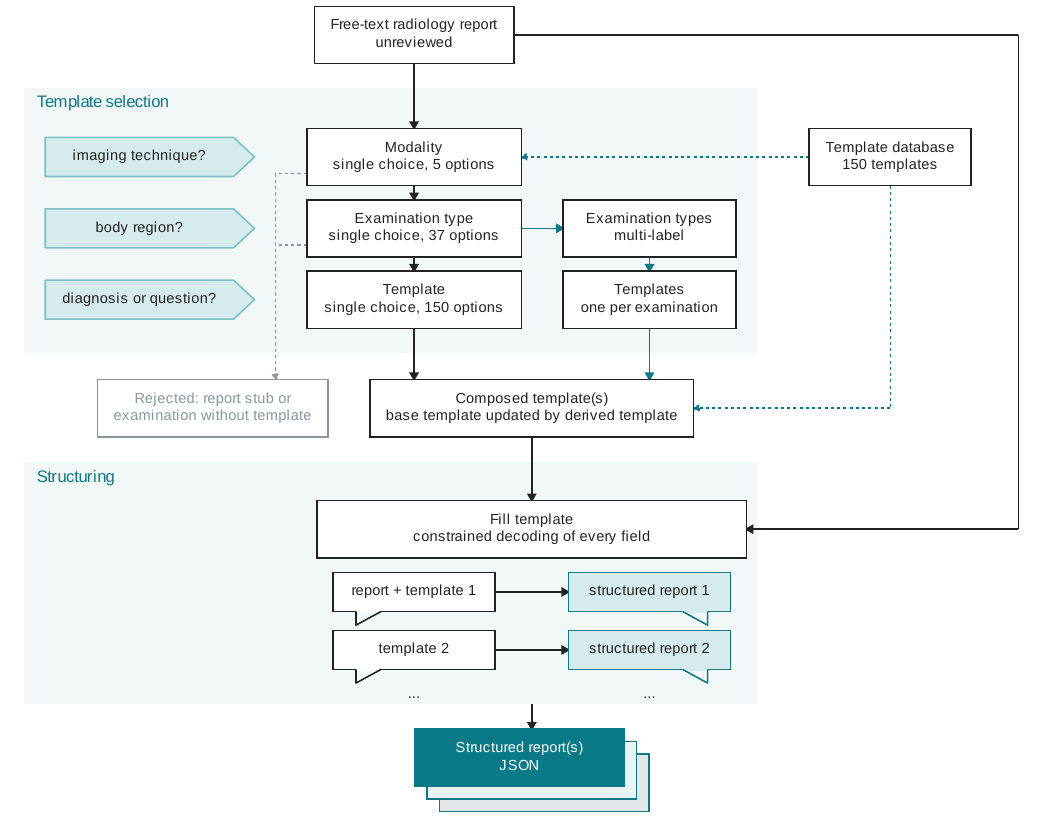}
\caption{Pipeline for autonomous structuring of free-text radiology reports. In the template selection stage the large language model answers three questions in constrained-decoding steps, imaging technique (modality), body region (examination type) and diagnosis or question (template), each narrowing the options of the next. Reports describing several examinations continue with multi-label selection of examination types and one template per examination (teal). Report stubs and examinations without a template are rejected (dashed grey). The template database supplies the options and the template files (dashed teal). The selected templates are composed from base and derived template and filled with the report content in the structuring stage, one call per template.}\label{fig:2}
\end{figure}

The pipeline takes unreviewed report text and proceeds in two stages (\hyperref[fig:2]{Figure~2}). In the selection stage the LLM classifies the report in three successive steps, modality, examination type and template, with each answer narrowing the options of the next. At every step two reserved responses reject report stubs and examinations without a template. If the report describes several examinations, the pipeline switches to multi-label classification of examination types and templates (\hyperref[fig:S2]{Figure~S2}). In the structuring stage the LLM fills the selected templates with the report content zero-shot, one template per call. Fields without information in the source report receive the placeholder ``not mentioned''.

Every response is produced with constrained decoding against a format generated at run time from the template index and the template files, the allowed options in the selection stage and the sections of the composed template in the structuring stage (Supplemental Methods S3). The pipeline is implemented in Python with LangChain 1.2.18 and runs against any OpenAI API compatible server.

\subsection{Model}

gpt-oss-120B \cite{gptoss}, a 4-bit quantized mixture-of-experts model with 117 billion total and 5.1 billion active parameters, was used for all experiments and for the archive. The context window was 15,000 tokens and reasoning effort was set to the lowest level. The model ran on one H100 GPU at Charité served through Ollama during development and on one H200 GPU at TUM served through vLLM \cite{ref18} for testing and archive processing.

\subsection{Test Datasets}

Two test sets were drawn from the TUM archive (\hyperref[fig:1]{Figure~1}). For template selection, 1,000 reports were sampled at random, 21 entries without a report identifier were removed, and a board-certified radiologist labeled the remaining 979 reports with modality, examination types and optimal templates or marked them as non-structurable (report stub or examination without a template). A second board-certified radiologist reviewed all labels (readers KKB and LCA). Forty-six reports combining two modalities were not scored, the 19 non-structurable reports served as the rejection test, and the remaining 914 reports formed the selection test set (766 single-region, 148 multi-region).

For structuring quality, 1,070 radiography and CT reports were sampled from the pipeline output so that all 70 templates of these modalities were represented by 9 to 25 reports. Five radiology residents (MF, ES, MK, LSchm, and LSchu; 2 to 3 years of experience) rated the selected template as appropriate, inappropriate or unclear, and two board-certified radiologists (KKB and LCA) adjudicated the 44 unclear ratings in consensus. For the 920 reports rated appropriate by the residents, the residents corrected the structured output field by field against the source report and flagged content not supported by the source as hallucinated. The corrected output of these 920 reports served as the reference standard (\hyperref[tab:1]{Table~1}). Labels and corrections were entered in software developed by the authors for this study.

\begin{table}[!htbp]
\centering
\caption{Template library and test datasets}\label{tab:1}
\begingroup
\setlength{\tabcolsep}{2.8pt}
\renewcommand{\arraystretch}{1.20}
\fontsize{8.5}{10.6}\selectfont
\hyphenpenalty=10000\exhyphenpenalty=10000
\begin{tabular}{@{}>{\RaggedRight\arraybackslash}p{31mm}>{\raggedleft\arraybackslash}p{9.0mm}@{\hspace{1.0mm}}>{\raggedleft\arraybackslash}p{5.6mm}@{\hspace{1.0mm}}>{\raggedleft\arraybackslash}p{4.6mm}@{\hspace{1.0mm}}>{\raggedleft\arraybackslash}p{4.8mm}@{\hspace{4mm}}>{\RaggedRight\arraybackslash}p{20mm}>{\raggedleft\arraybackslash}p{8.0mm}@{\hspace{1.0mm}}>{\raggedleft\arraybackslash}p{7.0mm}@{\hspace{1.0mm}}>{\raggedleft\arraybackslash}p{6.2mm}@{\hspace{4mm}}>{\raggedleft\arraybackslash}p{9.8mm}@{\hspace{1.0mm}}>{\raggedleft\arraybackslash}p{7.0mm}@{\hspace{1.0mm}}>{\raggedleft\arraybackslash}p{6.2mm}@{}}
\toprule
\textbf{Modality} & \multicolumn{4}{>{\RaggedRight\arraybackslash}p{27.0mm}}{\textbf{Templates (base, derived, independent)}} & \textbf{Examination types} & \multicolumn{3}{>{\RaggedRight\arraybackslash}p{23.2mm}}{\textbf{Selection test reports (single, multi)}} & \multicolumn{3}{>{\RaggedRight\arraybackslash}p{25.0mm}@{}}{\textbf{Reader study reviewed (appropriate, corrected)}} \\
\midrule
Radiography & 25 & (6, & 17, & 2) & 6 & 602 & (568, & 34) & 622 & (544, & 513) \\
\rowcolor{tabshade}
CT & 45 & (10, & 30, & 5) & 10 & 217 & (110, & 107) & 448 & (417, & 407) \\
MRI & 43 & (6, & 24, & 13) & 11 & 59 & (52, & 7) & \textemdash &  &  \\
\rowcolor{tabshade}
Ultrasound & 30 & (0, & 0, & 30) & 6 & 26 & (26, & 0) & \textemdash &  &  \\
Interventional radiology & 7 & (2, & 4, & 1) & 4 & 10 & (10, & 0) & \textemdash &  &  \\
\rowcolor{tabshade}
\midrule
\textbf{Total} & \textbf{150} & \textbf{(24,} & \textbf{75,} & \textbf{51)} & \textbf{37} & \textbf{914} & \textbf{(766,} & \textbf{148)} & \textbf{1,070} & \textbf{(961,} & \textbf{920)} \\
\bottomrule
\end{tabular}
\endgroup
\par\vspace{3pt}{\footnotesize\RaggedRight Creation of both test sets and the excluded reports in \hyperref[fig:1]{Figure~1}. Templates listed under several examination types are counted once, the pain-therapy template shared by CT and interventional radiology under CT. The 31 CT combination types are not templates. Appropriate, template rated appropriate by the residents or, for 44 unclear ratings, by two adjudicating board-certified radiologists in consensus. Corrected, reports rated appropriate by the residents and corrected field by field, the reference for the field-level scores. Median report length was 92 words for radiography, 177 for CT, 174 for MRI, 149 for ultrasound and 220 for interventional radiology.\par}
\end{table}

\subsection{Evaluation Metrics and Statistical Analysis}

Template selection was evaluated per report with all 914 reports of the selection test set as denominator. Sensitivity is the fraction that received at least one template, specificity the fraction of the 19 non-structurable reports that were rejected. A report was optimal if the selected template set matched the expert label and appropriate if every selected template shared its base template with the labeled one. Template identifiers were compared after normalization of spelling, which affected two labels. Results are stratified by modality and by single-region versus multi-region reports. Two ablations supplied the labeled modality to the pipeline and raised the reasoning effort to medium and high on the 111 selection test reports that the pipeline routed to multi-label classification, paired differences were tested with two-sided exact McNemar tests.

Structuring quality was scored field by field against the corrected reference, so the template text itself does not contribute. Semantic agreement was measured as semantic textual similarity (STS), the cosine similarity of Llama-Embed-Nemotron-8B embeddings \cite{ref19}, lexical agreement as BLEU \cite{ref20}, both macro-averaged over templates and additionally over content fields only, fields with report content in output or reference. A field counted as correct when the residents left it unchanged, whether it held report content or the placeholder for absent information, and as edited otherwise. The hallucination rate is the fraction of corrected reports with at least one flagged field. A board-certified radiologist classified the primary error of the 50 lowest-STS reports per modality as incomplete transcription, wrong or misplaced transcription, or hallucination.

Proportions are reported with Wilson 95\% confidence intervals (CIs). CIs for STS and BLEU were computed with a percentile bootstrap of reports within templates (10,000 resamples). Analyses were performed in Python 3.12 with SciPy. P < .05 indicated significance.

\subsection{Archive Processing}

The complete TUM report archive was processed with the same pipeline and model without human oversight. Rows without a valid report identifier and header rows of examinations imported from other institutions were excluded before processing. A report counted as structured when the pipeline returned at least one template with valid JSON output. Throughput was derived from the processing logs on one H200 GPU and, for one report at a time, measured in a serial benchmark of 100 reports run twice.

\section{Results}

The selection test set comprised 914 reports drawn at random from the TUM archive (766 single-region, 148 multi-region), the structuring test set the 920 corrected reports (513 radiography, 407 CT, 70 templates) (\hyperref[fig:1]{Figure~1}, \hyperref[tab:1]{Table~1}).

\subsection{Template Selection}

The pipeline assigned at least one template to 869 of 914 reports (95.1\%, 95\% CI 93.5 to 96.3). Of the 19 non-structurable reports (\hyperref[fig:1]{Figure~1A}), 18 were rejected (94.7\%, 75.4 to 99.1). Sensitivity ranged from 65.4\% for ultrasound (17 of 26) to 97.7\% for CT (212 of 217) (\hyperref[tab:2]{Table~2}).

\begin{figure}[!htbp]
\centering
\includegraphics[width=\linewidth]{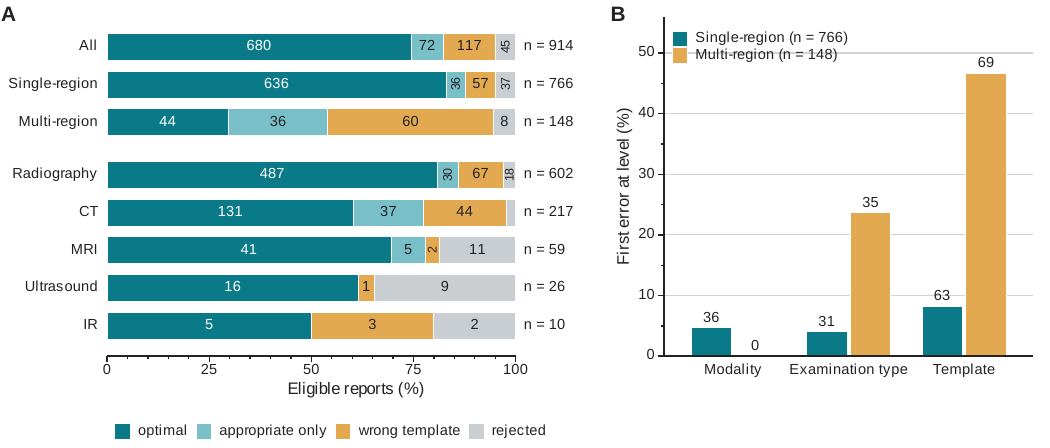}
\caption{Template selection on 914 eligible external test reports. A, outcome of template selection per report as percentage of eligible reports, for all reports, by number of described regions and by modality, with report counts inside the segments and n at the bar end. B, level of the hierarchy at which the first error occurred, as percentage of the group, with report counts above the bars. IR, interventional radiology.}\label{fig:3}
\end{figure}

The optimal template set was selected for 680 of 914 reports (74.4\%, 71.5 to 77.1) and an appropriate set for 752 (82.3\%, 79.7 to 84.6) (\hyperref[tab:2]{Table~2}, \hyperref[fig:3]{Figure~3A}). Single-region reports reached 636 of 766 optimal (83.0\%, 80.2 to 85.5) and 672 of 766 appropriate (87.7\%, 85.2 to 89.9). Multi-region reports reached 44 of 148 optimal (29.7\%, 23.0 to 37.5) and 80 of 148 appropriate (54.1\%, 46.0 to 61.9).

\begin{table}[!htbp]
\centering
\caption{Template selection on the 914 reports of the selection test set}\label{tab:2}
\begingroup
\setlength{\tabcolsep}{2.8pt}
\renewcommand{\arraystretch}{1.20}
\fontsize{8.5}{10.6}\selectfont
\begin{tabular}{@{}>{\RaggedRight\arraybackslash}p{26.59mm}>{\RaggedRight\arraybackslash}p{9.14mm}>{\centering\arraybackslash}p{34.06mm}>{\centering\arraybackslash}p{34.06mm}>{\centering\arraybackslash}p{34.06mm}>{\centering\arraybackslash}p{28.25mm}@{}}
\toprule
\textbf{Group} & \textbf{n} & \textbf{Structured, \% (95\% CI)} & \textbf{Optimal, \% (95\% CI)} & \textbf{Appropriate, \% (95\% CI)} & \textbf{Optimal given structured, \%} \\
\midrule
All & 914 & 95.1 (93.5, 96.3) & 74.4 (71.5, 77.1) & 82.3 (79.7, 84.6) & 78.3 \\
\rowcolor{tabshade}
Single-region & 766 & 95.2 (93.4, 96.5) & 83.0 (80.2, 85.5) & 87.7 (85.2, 89.9) & 87.2 \\
Multi-region & 148 & 94.6 (89.7, 97.2) & 29.7 (23.0, 37.5) & 54.1 (46.0, 61.9) & 31.4 \\
\rowcolor{tabshade}
Radiography & 602 & 97.0 (95.3, 98.1) & 80.9 (77.6, 83.8) & 85.9 (82.9, 88.4) & 83.4 \\
CT & 217 & 97.7 (94.7, 99.0) & 60.4 (53.7, 66.6) & 77.4 (71.4, 82.5) & 61.8 \\
\rowcolor{tabshade}
MRI & 59 & 81.4 (69.6, 89.3) & 69.5 (56.9, 79.7) & 78.0 (65.9, 86.6) & 85.4 \\
Ultrasound & 26 & 65.4 (46.2, 80.6) & 61.5 (42.5, 77.6) & 61.5 (42.5, 77.6) & 94.1 \\
\rowcolor{tabshade}
Interventional radiology & 10 & 80.0 (49.0, 94.3) & 50.0 (23.7, 76.3) & 50.0 (23.7, 76.3) & 62.5 \\
\bottomrule
\end{tabular}
\endgroup
\par\vspace{3pt}{\footnotesize\RaggedRight Structured, at least one template assigned. Optimal, selected template set identical to the expert label. Appropriate, every selected template shares its base template with the labeled one (includes optimal). Denominator for the first three rates is all eligible reports of the group. Specificity on the 19 non-structurable reports was 94.7\% (18 of 19, 75.4 to 99.1).\par}
\end{table}

Among single-region reports, the first error occurred at the modality step in 36 reports, at the examination-type step in 31 and at the template step in 63. Among multi-region reports, 35 first errors occurred at the examination-type step and 69 at the template step (\hyperref[fig:3]{Figure~3B}). Of the 57 single-region reports with a template outside the appropriate family, 39 involved the skeletal radiography templates, most often a postoperative extremity radiograph assigned to a pelvis template (\hyperref[tab:S3]{Table~S3}). Per-template results are given in \hyperref[tab:S2]{Table~S2}. Supplying the labeled modality raised optimal selection by 19 of 914 reports (2.1 percentage points, 0.1 to 4.0, P = .05), and medium or high reasoning effort on these 111 reports did not change optimal selection (45, 47 and 41 reports, P = .79 and P = .48) (Tables S4 and S5).

\subsection{Structuring Quality}

Among the reviewed reports (\hyperref[fig:1]{Figure~1B}), the residents rated the selected template appropriate for 87.5\% of radiography reports (544 of 622, 84.6 to 89.8) and 93.1\% of CT reports (417 of 448, 90.3 to 95.1), including 41 of 44 unclear ratings that two board-certified radiologists adjudicated in consensus (\hyperref[fig:5]{Figure~5A}).

\begin{figure}[!htbp]
\centering
\includegraphics[width=\linewidth]{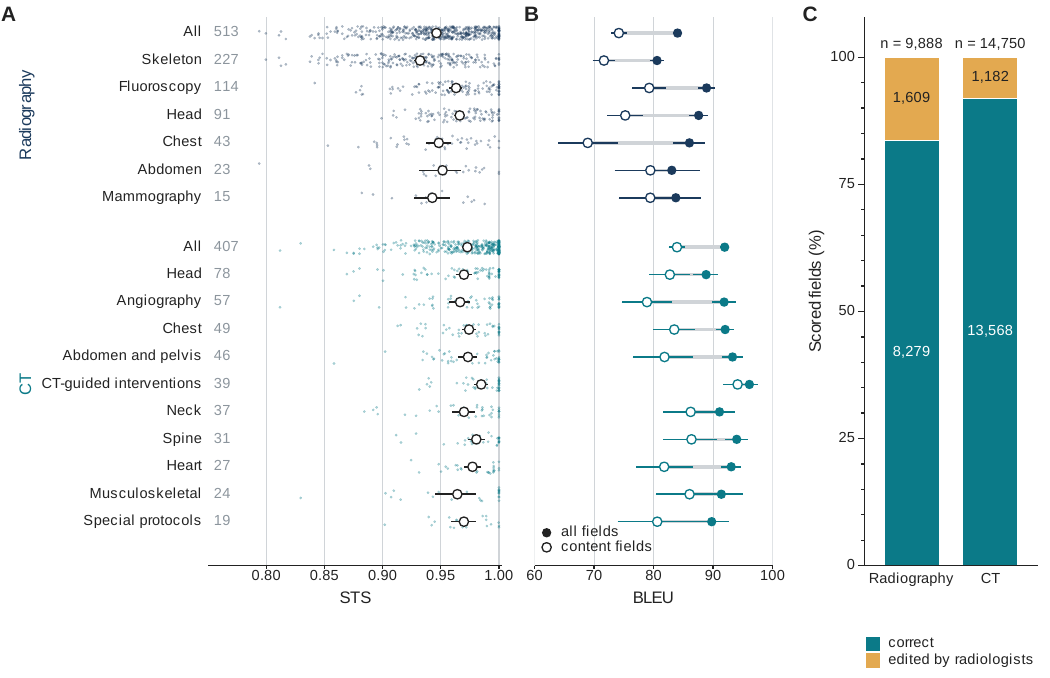}
\caption{Structuring quality against the resident-corrected reference for 513 radiography and 407 CT reports. A, STS per report (points) and macro mean with 95\% CI (open circle with line) by examination type, report counts in grey. B, macro BLEU over all fields (filled) and over fields with content in output or reference (open) with 95\% CI. C, share of the 24,638 scored fields the radiologist left unchanged (correct) or edited, with field counts inside the bars. STS, semantic textual similarity. BLEU, bilingual evaluation understudy.}\label{fig:4}
\end{figure}

Macro STS between pipeline output and corrected reference was 0.95 (0.94 to 0.95) for radiography and 0.97 (0.97 to 0.98) for CT. Macro BLEU was 84.0 (83.3 to 84.8) and 91.9 (91.2 to 92.6) (\hyperref[fig:4]{Figure~4A}, B, \hyperref[tab:3]{Table~3}). Macro STS by examination type ranged from 0.93 to 0.98 (\hyperref[tab:S1]{Table~S1}). Of 24,638 scored fields, the residents left 21,847 (88.7\%) unchanged, 8,279 of 9,888 radiography fields (83.7\%) and 13,568 of 14,750 CT fields (92.0\%), and edited 1,609 and 1,182 (\hyperref[fig:4]{Figure~4C}). Restricted to fields with content in output or reference, macro BLEU was 74.2 (72.9 to 75.4) for radiography and 83.9 (82.5 to 85.3) for CT. Output and corrected reference were identical in every field for 7.6\% of radiography and 25.6\% of CT reports (\hyperref[tab:3]{Table~3}).

\begin{table}[!htbp]
\centering
\caption{Structuring quality against the resident-corrected reference}\label{tab:3}
\begingroup
\setlength{\tabcolsep}{2.8pt}
\renewcommand{\arraystretch}{1.20}
\fontsize{8.5}{10.6}\selectfont
\begin{tabular}{@{}>{\RaggedRight\arraybackslash}p{22.17mm}>{\RaggedRight\arraybackslash}p{12.31mm}>{\centering\arraybackslash}p{13.14mm}>{\centering\arraybackslash}p{31.20mm}>{\centering\arraybackslash}p{31.20mm}>{\centering\arraybackslash}p{31.20mm}>{\centering\arraybackslash}p{22.99mm}@{}}
\toprule
\textbf{Modality} & \textbf{Reports} & \textbf{Templates} & \textbf{STS (95\% CI)} & \textbf{BLEU, all fields (95\% CI)} & \textbf{BLEU, content fields (95\% CI)} & \textbf{Identical reports, n (\%)} \\
\midrule
Radiography & 513 & 25 & 0.95 (0.94, 0.95) & 84.0 (83.3, 84.8) & 74.2 (72.9, 75.4) & 39 (7.6) \\
\rowcolor{tabshade}
CT & 407 & 45 & 0.97 (0.97, 0.98) & 91.9 (91.2, 92.6) & 83.9 (82.5, 85.3) & 104 (25.6) \\
\bottomrule
\end{tabular}
\endgroup
\par\vspace{3pt}{\footnotesize\RaggedRight Macro scores weight templates equally, CIs from a percentile bootstrap of reports within templates. Content fields, fields with content in output or reference (the placeholder for information absent from the report excluded on both sides). Identical, output and corrected reference equal in every field. Results by examination type in \hyperref[tab:S1]{Table~S1}.\par}
\end{table}

\subsection{Hallucinations and Error Types}

\begin{figure}[!htbp]
\centering
\includegraphics[width=\linewidth]{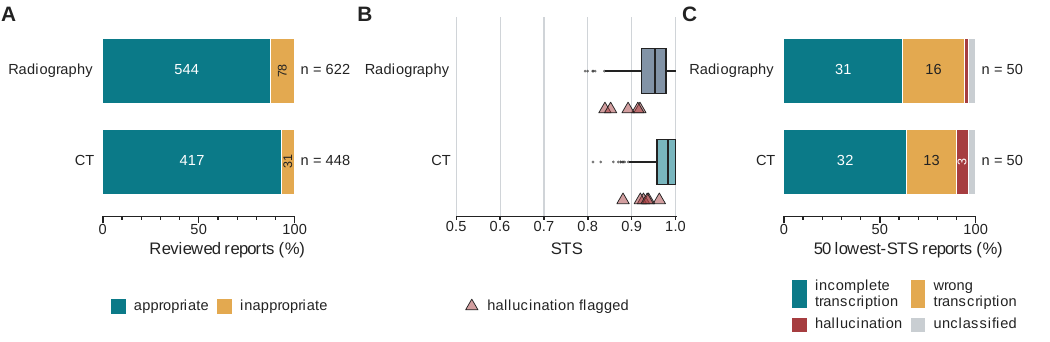}
\caption{Template ratings, hallucination flags and error types in the reader study. A, rating of the selected template for 622 radiography and 448 CT reports after adjudication of the unclear ratings, with report counts inside the segments. B, STS of the 513 and 407 corrected reports as box plots (median, interquartile range, 1.5 IQR whiskers, outliers as points) with the reports flagged for unsupported content marked as triangles. C, primary error type in the 50 lowest-STS reports per modality, with report counts inside the segments where they fit (hallucination 1 and unclassified 2 in radiography, unclassified 2 in CT).}\label{fig:5}
\end{figure}

Residents flagged content not supported by the source report in 5 of 513 radiography reports (1.0\%, 0.4 to 2.3) and 6 of 407 CT reports (1.5\%, 0.7 to 3.2). All 11 flagged reports lay below the 30th STS percentile of their modality (\hyperref[fig:5]{Figure~5B}). In the 50 lowest-STS reports per modality, the primary error was incomplete transcription in 31 radiography and 32 CT reports, wrong or misplaced transcription in 16 and 13, and hallucination in 1 and 3, with 2 reports per modality unclassified (\hyperref[fig:5]{Figure~5C}).

\subsection{Archive-Scale Processing}

\begin{figure}[!htbp]
\centering
\includegraphics[width=\linewidth]{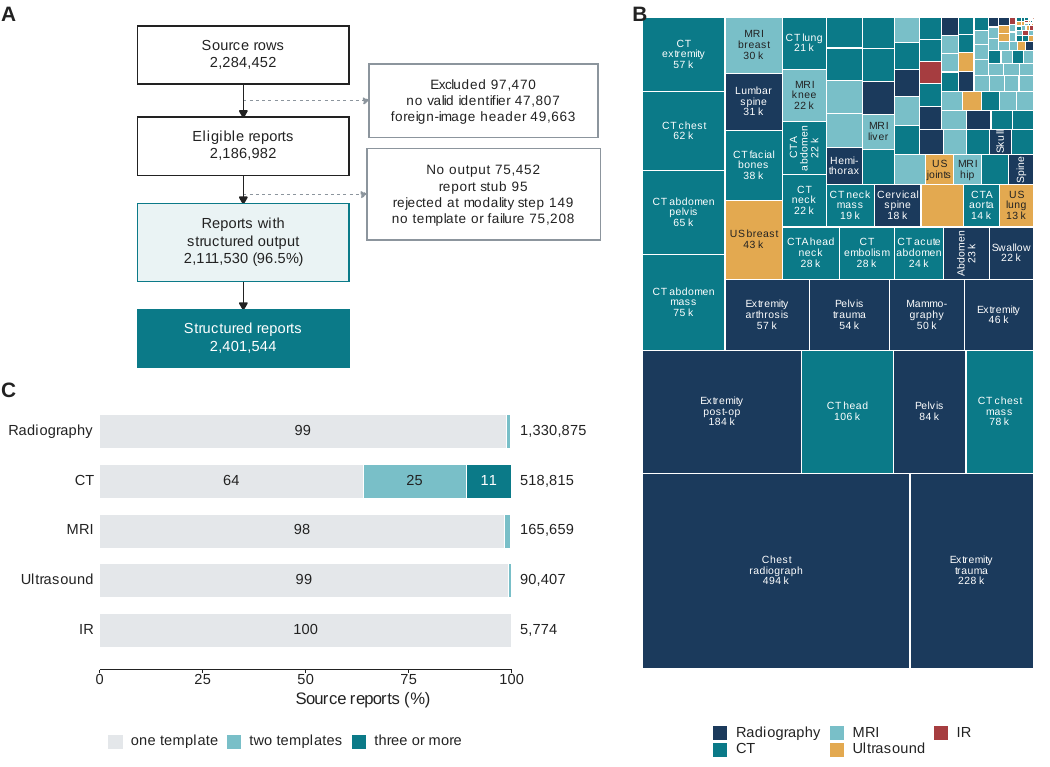}
\caption{Structuring of the complete TUM report archive. A, flow from source rows to structured reports. B, structured reports per template, area proportional to count, color by modality, templates labeled where the tile holds the name, with counts in thousands where space allows (all templates in \hyperref[tab:S6]{Table~S6}). C, number of templates per source report by modality. IR, interventional radiology.}\label{fig:6}
\end{figure}

The TUM archive held 2,284,452 report rows. After exclusion of 47,807 rows without a valid identifier and 49,663 foreign-image headers, 2,186,982 reports were eligible. The pipeline produced structured output for 2,111,530 of them (96.5\%) as 2,401,544 structured reports (\hyperref[tab:4]{Table~4}, \hyperref[fig:6]{Figure~6A}). The remaining 75,452 reports had no output (\hyperref[tab:4]{Table~4}). All structured reports were valid JSON, followed the field structure of their template and were linked to the source report by identical text.

Of the 142 distinct templates in use, the chest radiograph template accounted for 493,844 structured reports, the trauma and postoperative extremity templates for 227,864 and 183,504, and the CT head template for 106,061 (\hyperref[fig:6]{Figure~6B}, \hyperref[tab:S6]{Table~S6}). Radiography, MRI and ultrasound reports received one template in 98 to 99\% of cases. CT reports received two templates in 25\% and three or more in 11\% of cases (combination types, \hyperref[fig:6]{Figure~6C}, \hyperref[fig:S1]{Figure~S1}).

The pipeline was able to structure 1,258 reports per hour on one H200 GPU, about 1,680 GPU hours for the 2,111,530 structured reports. In a serial benchmark, one report took a median of 10.6 s (IQR 9.1 to 13.6) with a median of 2,737 input and 1,431 generated tokens.

\begin{table}[!htbp]
\centering
\caption{Structured archive by modality}\label{tab:4}
\begingroup
\setlength{\tabcolsep}{2.8pt}
\renewcommand{\arraystretch}{1.20}
\fontsize{8.5}{10.6}\selectfont
\begin{tabular}{@{}>{\RaggedRight\arraybackslash}p{30.26mm}>{\RaggedRight\arraybackslash}p{35.31mm}>{\RaggedRight\arraybackslash}p{26.90mm}>{\RaggedRight\arraybackslash}p{19.33mm}>{\RaggedRight\arraybackslash}p{56.32mm}@{}}
\toprule
\textbf{Modality} & \textbf{Reports with structured output} & \textbf{Structured reports} & \textbf{Templates in use} & \textbf{Largest template (share of structured reports)} \\
\midrule
Radiography & 1,330,875 & 1,354,491 & 25 & chest radiograph (36.5\%) \\
\rowcolor{tabshade}
CT & 518,815 & 780,857 & 45 & CT head (13.6\%) \\
MRI & 165,659 & 169,097 & 38 & MRI breast (17.9\%) \\
\rowcolor{tabshade}
Ultrasound & 90,407 & 91,325 & 27 & breast (47.2\%) \\
Interventional radiology & 5,774 & 5,774 & 8 & image-guided interventions (82.4\%) \\
\rowcolor{tabshade}
\midrule
\textbf{Total} & \textbf{2,111,530} & \textbf{2,401,544} & \textbf{142} & \textbf{\textemdash} \\
\bottomrule
\end{tabular}
\endgroup
\par\vspace{3pt}{\footnotesize\RaggedRight 2,284,452 source rows, 47,807 without valid identifier, 49,663 foreign-image headers, 2,186,982 eligible reports, 96.5\% with structured output. Without output 75,452 (95 report stubs, 149 rejected at the modality step, 75,208 without template or with a technical failure). The pain-therapy template exists under CT and interventional radiology, so 143 template folders correspond to 142 templates. Throughput 1,258 structured reports per hour on one H200 GPU.\par}
\end{table}

\FloatBarrier

\section{Discussion}

An open-weight LLM converted the complete report archive of a university hospital, 2.1 million free-text reports from five modalities and sixteen years, into structured reports on local GPU infrastructure and without a human in the loop. The pipeline chose among 150 templates on its own, rejected reports that fit none, and filled the chosen templates with content that radiology at the second center left largely unchanged.

At 1,258 reports per hour on one H200 GPU, the archive corresponds to about 1,680 GPU hours. The same tokens sent to a commercial API at the mid-tier list price of 2 USD/12 USD per million input/output tokens would cost about 48,000 USD, and about 120,000 USD at the flagship tier. Local deployment has been advocated mainly for privacy \cite{ref21},  and studies comparing open-weight with proprietary models on radiology text found similar accuracy \cite{ref22,ref23,ref24}. At archive scale, cost and privacy both favor local deployment.

LLMs are sensitive to the number and arrangement of answer options, and classification accuracy falls as option lists grow \cite{ref25}. Supplying the labeled modality raised optimal selection by 2 percentage points, so archive metadata would add little. Remaining errors occur at the template step and in multi-region reports. Each template is built around one primary disease, but a patient with several diseases in one examination has no single correct template, and the pipeline must return a set of templates rather than one. Such reports account for 16\% of the selection test set but for 44\% of the selection errors, and template selection for them is the main open problem of the design.

Common data elements were proposed to make report content exchangeable between systems by defining each element once and reusing it across templates \cite{ref26}, and the European Society of Radiology restated in 2023 that comparability across reports is the purpose of structured reporting \cite{ref4}. Base and derived templates implement this principle in a form the LLM can fill. Every CT chest report, whether written for a mass, an embolism or a routine question, carries the same base sections, so an archive query addresses one field name. Schema-stratified report supervision has been shown to produce image representations with separated concepts and better visual grounding than free-text supervision \cite{ref27}.

In a clinician-annotated study of note generation, omissions (3.45\% of sentences) exceeded hallucinations (1.47\%) \cite{ref28}, and hallucination rates across medical summarization studies range from 1.5\% to 62\% depending on task and definition \cite{ref29}. In this study, radiologists flagged unsupported content in 1.0\% and 1.5\% of corrected reports, and the primary error in the lowest-scoring reports was incomplete transcription in about two thirds of cases. A missing statement is visible as an empty or placeholder field, invented content is not, so omission is the more tractable failure. All flagged reports lay below the 30th STS percentile of their modality, so an embedding-based similarity between output and source can direct human review to the reports that need it.

This study has limitations. The reports were German, and open-weight models score higher in English than in other languages on medical tasks \cite{ref30}. Woźnicki et al. reached a Matthews correlation coefficient of 0.75 for English against 0.66 for German chest radiograph reports with Llama-2 \cite{ref13}, so the results here are likely a lower bound for English archives. Structuring quality was assessed for radiography and CT only. CT had the lowest optimal selection rate, so the reader study covered the weakest modality, but MRI, ultrasound and interventional radiology remain untested at field level. The selection test set of 914 labeled reports spreads over 150 templates, and 60 templates had fewer than 10 test reports, so per-template accuracy is imprecise. One model was used at low reasoning effort. The pipeline is model agnostic, and smaller open-weight models released since, among them Gemma 4 and Qwen3.5, are reported to reach comparable benchmark accuracy with fewer parameters, so throughput and cost are likely to improve.

\begin{samepage}
A department can turn its entire report archive into structured, queryable data with an open-weight LLM on its own hardware. The structured archive makes sixteen years of imaging findings available for cohort identification, quality control and the training of imaging models, and the template hierarchy allows other institutions to adapt the templates rather than the method.

\end{samepage}

\FloatBarrier\section{Declarations}

\subsection{Funding}
This research is funded by the European Union under HORIZON-HLTH-2022-TOOL-12-two-stage (COMFORT; grant agreement No. 101079894) and HORIZON-HLTH-2025-01 (TWIN-X; grant agreement No. 101287679; topic HORIZON-HLTH-2025-01-TOOL-03). Views and opinions expressed are however those of the author(s) only and do not necessarily reflect those of the European Union or the European Health and Digital Executive Agency (HADEA). Neither the European Union nor the granting authority can be held responsible for them. The funder had no role in study design, data collection, analysis, or writing.

\subsection{Conflicts of interest}
JN reports research collaboration agreements with Briya Labs Ltd., Cercare Medical, and mediaire, and has received grant funding from the Berlin Institute of Health (BIH) and the European Union (COMFORT, project number 101079894). KKB reports speaker fees for Canon Medical Systems, consultant fees for Novartis, and grant funding from the European Union (COMFORT, project number 101079894; TWIN-X, project number 101287679). RC, AB, and AL report grant funding from the European Union (COMFORT, project number 101079894). LCA reports grant funding from the European Union (COMFORT, project number 101079894; TWIN-X, project number 101287679). CIB reports grant funding from the European Union (TWIN-X, project number 101287679).

\subsection{Data and code availability}
The report texts cannot be shared because they contain protected health information. The complete pipeline, all structuring templates with their German-language definitions and the template index are available at \url{https://github.com/fputtkammer/RadFields}.

\subsection{Use of artificial intelligence}
A generative artificial intelligence assistant (Claude, Anthropic) was used under author supervision for drafting and editing text, for the analysis and figure code and for the assembly of the supplement. Codex (OpenAI) was used to prepare the LaTeX preprint. The authors reviewed and verified all content and take full responsibility for it.

\subsection{Correspondence}
Keno K. Bressem, MD; Institute for Artificial Intelligence in Medicine, University Hospital Essen, Girardetstraße 2, 45131 Essen, Germany; keno.bressem@nct-west.de

\subsection{Abbreviations}
LLM = large language model, STS = semantic textual similarity, BLEU = bilingual evaluation understudy, CI = confidence interval, IQR = interquartile range, GPU = graphics processing unit, API = application programming interface

\clearpage\section{Supplementary material}
\setcounter{figure}{0}\renewcommand{\thefigure}{S\arabic{figure}}
\setcounter{table}{0}\renewcommand{\thetable}{S\arabic{table}}
\subsection{Supplemental Methods S1. Template development and format}

To determine which examination types and diagnoses the templates must cover, 10,000 reports were drawn at random from the Charité archive and screened with gpt-oss-120B in three steps per report. First, the LLM assigned the imaging modality from a preset list. Second, it assigned the examined body region or organ system from a list of examples that grew during screening, as the LLM could add entries not yet on the list. Third, it extracted the ICD-10 code of the main diagnosis, or generated one if the report did not state a code. A board-certified radiologist merged redundant body region categories, and the frequency of three-digit ICD-10 categories was tabulated per modality and body region. Two board-certified radiologists reviewed these tables and defined a base template for each common examination type and derived templates for diagnosis categories exceeding 5\% of reports within an examination type.

Each template is a YAML file in which every key is a section of the structured report and the value defines its type. A string defines a free-text section and describes the expected content to the LLM, a mapping defines subsections, a list of strings defines several free-text fields, and a list of mappings defines a repeatable section, for example one entry per lesion. Section titles are valid Python identifiers, so a Pydantic model is generated from each template at run time and serves as the response format for constrained decoding. Templates are written in German. Fields without information in the source report are filled with the placeholder “not mentioned”, which separates missing information from omitted content.

\subsection{Supplemental Methods S2. Template index, inheritance, and combination types}

The template index is a YAML file that lists, per modality and examination type, every template with a unique name, a German description shown to the LLM during template selection, the path of its template file, and its position in the hierarchy. A base template carries the field children with the names of its derived templates, a derived template carries the field parent with the name of its base template, and an independent template carries neither. The description is the only information the LLM sees at the template step, so descriptions state the clinical question a template answers, for example pulmonary embolism or tumor staging for CT chest.

At run time the pipeline composes the final template by loading the base template and updating it with the derived template. Sections present only in the base template or only in the derived template are kept, sections present in both take the definition of the derived template. A change to a base template therefore reaches all its derived templates, and structured reports of one examination type share the sections of the base template regardless of the clinical question. The hierarchy has two levels.

Combination types are defined in a separate file as sets of examination types that occur together in one CT report, for example CT chest with CT abdomen and pelvis for tumor staging. The 31 combination types are defined for CT only. For a report of a combination type, the LLM selects one template per constituent examination type from the templates of these examination types, and the report is structured once per selected template. Templates and combination types can be added, changed, or removed by editing the index and the template files without a change to the pipeline code.

\subsection{Supplemental Methods S3. Template selection pathways and constrained decoding}

Every classification step is a constrained decoding call. At run time the pipeline builds a response format that lists the allowed options of the step together with reserved responses, “no findings” for report stubs, “none of the listed options” for examinations without a template, and, at the examination type step, “several examinations described”. The LLM can only return an item of this list. A response that does not parse against the format is regenerated up to three times.

\hyperref[fig:S2]{Figure~S2} shows the pathways. A basic report, which describes one examination, passes through single-choice classification of modality, examination type, and template. At the template step, the LLM sees the names and descriptions of all templates of the selected examination type. A multi-exam report, in which several examinations of one modality are described in separate paragraphs, for example radiographs of both hands and both feet, triggers the reserved response “several examinations described” at the examination type step. The pipeline then continues with multi-label classification, first of examination types, then of one template per selected examination type. A combined CT report, in which several examination types are described in one continuous text, typically a staging examination, is assigned a combination type at the examination type step and continues with multi-label template selection constrained to one template per constituent examination type (Supplemental Methods S2). The two rejection responses end the pipeline at any step.

In the structuring stage, the response format is the Pydantic model generated from the composed template (Supplemental Methods S1). The LLM receives the report text and the model and fills one template per call. Reports with several templates are structured in successive calls with the same report text. Calls are independent, so the pipeline carries no state between reports.

\subsection{Supplemental tables}
\begin{table}[!htbp]
\centering
\caption{Structuring quality by examination type}\label{tab:S1}
\begingroup
\setlength{\tabcolsep}{2.8pt}
\renewcommand{\arraystretch}{1.20}
\fontsize{8.0}{10.1}\selectfont
\begin{tabular}{@{}>{\RaggedRight\arraybackslash}p{17.84mm}>{\RaggedRight\arraybackslash}p{21.09mm}>{\centering\arraybackslash}p{10.54mm}>{\centering\arraybackslash}p{13.79mm}>{\centering\arraybackslash}p{25.96mm}>{\centering\arraybackslash}p{25.96mm}>{\centering\arraybackslash}p{25.96mm}>{\centering\arraybackslash}p{21.09mm}@{}}
\toprule
\textbf{Modality} & \textbf{Examination type} & \textbf{Reports} & \textbf{Templates} & \textbf{STS (95\% CI)} & \textbf{BLEU, all fields (95\% CI)} & \textbf{BLEU, content fields (95\% CI)} & \textbf{Identical reports, n (\%)} \\
\midrule
Radiography & All & 513 & 25 & 0.95 (0.94, 0.95) & 84.0 (83.3, 84.8) & 74.2 (72.9, 75.4) & 39 (7.6) \\
\rowcolor{tabshade}
Radiography & Abdomen & 23 & 1 & 0.95 (0.93, 0.97) & 83.1 (78.0, 87.8) & 79.5 (73.4, 84.8) & 2 (8.7) \\
Radiography & Chest & 43 & 2 & 0.95 (0.94, 0.96) & 86.0 (83.2, 88.7) & 68.9 (64.0, 74.0) & 2 (4.7) \\
\rowcolor{tabshade}
Radiography & Fluoroscopy & 114 & 5 & 0.96 (0.96, 0.97) & 88.9 (87.5, 90.3) & 79.3 (76.3, 82.0) & 18 (15.8) \\
Radiography & Head & 91 & 4 & 0.97 (0.96, 0.97) & 87.6 (86.0, 89.1) & 75.2 (72.2, 78.2) & 10 (11.0) \\
\rowcolor{tabshade}
Radiography & Mammography & 15 & 1 & 0.94 (0.93, 0.96) & 83.7 (79.3, 88.0) & 79.4 (74.1, 84.5) & 0 (0.0) \\
Radiography & Skeleton & 227 & 12 & 0.93 (0.93, 0.94) & 80.6 (79.4, 81.8) & 71.7 (69.8, 73.6) & 7 (3.1) \\
\rowcolor{tabshade}
CT & All & 407 & 45 & 0.97 (0.97, 0.98) & 91.9 (91.2, 92.6) & 83.9 (82.5, 85.3) & 104 (25.6) \\
CT & Abdomen and pelvis & 46 & 5 & 0.97 (0.96, 0.98) & 93.3 (91.4, 95.0) & 81.8 (76.5, 86.6) & 9 (19.6) \\
\rowcolor{tabshade}
CT & Angiography & 57 & 6 & 0.97 (0.96, 0.98) & 91.8 (89.8, 93.8) & 78.9 (74.6, 83.1) & 11 (19.3) \\
CT & CT-guided interventions & 39 & 4 & 0.98 (0.98, 0.99) & 96.1 (94.5, 97.6) & 94.1 (91.7, 96.3) & 18 (46.2) \\
\rowcolor{tabshade}
CT & Chest & 49 & 6 & 0.97 (0.97, 0.98) & 92.0 (90.5, 93.5) & 83.5 (79.9, 87.0) & 12 (24.5) \\
CT & Head & 78 & 8 & 0.97 (0.96, 0.98) & 88.8 (86.6, 90.9) & 82.7 (79.3, 86.2) & 21 (26.9) \\
\rowcolor{tabshade}
CT & Heart & 27 & 3 & 0.98 (0.97, 0.98) & 93.0 (91.4, 94.7) & 81.8 (77.0, 86.6) & 3 (11.1) \\
CT & Musculoskeletal & 24 & 3 & 0.96 (0.95, 0.98) & 91.4 (87.4, 95.0) & 86.0 (80.3, 91.5) & 11 (45.8) \\
\rowcolor{tabshade}
CT & Neck & 37 & 4 & 0.97 (0.96, 0.98) & 91.1 (88.3, 93.7) & 86.2 (81.6, 90.6) & 8 (21.6) \\
CT & Special protocols & 19 & 2 & 0.97 (0.96, 0.98) & 89.8 (86.7, 92.6) & 80.6 (73.9, 87.0) & 2 (10.5) \\
\rowcolor{tabshade}
CT & Spine & 31 & 4 & 0.98 (0.97, 0.99) & 94.0 (92.0, 95.8) & 86.4 (81.6, 90.7) & 9 (29.0) \\
\bottomrule
\end{tabular}
\endgroup
\par\vspace{3pt}{\footnotesize\RaggedRight Macro STS and BLEU against the corrected reference for the 513 radiography and 407 CT reports of the structuring test set, by examination type. Macro scores weight templates equally, CIs from a percentile bootstrap of reports within templates (10,000 resamples). Content fields, fields with content in output or reference. Identical, output and corrected reference equal in every field. STS, semantic textual similarity. BLEU, bilingual evaluation understudy.\par}
\end{table}

\FloatBarrier
\begingroup
\setlength{\tabcolsep}{2.8pt}
\renewcommand{\arraystretch}{1.20}
\fontsize{8.3}{10.4}\selectfont
\begin{longtable}{@{}>{\RaggedRight\arraybackslash}p{21.34mm}>{\RaggedRight\arraybackslash}p{65.68mm}>{\centering\arraybackslash}p{15.43mm}>{\centering\arraybackslash}p{15.43mm}>{\centering\arraybackslash}p{15.43mm}>{\centering\arraybackslash}p{15.43mm}>{\centering\arraybackslash}p{15.43mm}@{}}
\caption{Template selection per labeled template}\label{tab:S2}\\
\toprule
\textbf{Modality} & \textbf{Labeled template} & \textbf{Test reports} & \textbf{Structured} & \textbf{Optimal} & \textbf{Appropriate} & \textbf{Optimal, \%} \\
\midrule
\endfirsthead
\multicolumn{7}{@{}l}{\sffamily\small\textbf{Table S2.} Template selection per labeled template (continued)}\\[4pt]
\toprule
\textbf{Modality} & \textbf{Labeled template} & \textbf{Test reports} & \textbf{Structured} & \textbf{Optimal} & \textbf{Appropriate} & \textbf{Optimal, \%} \\
\midrule
\endhead
\midrule\multicolumn{7}{r@{}}{\footnotesize Continued on next page}\\
\endfoot
\bottomrule
\endlastfoot
Radiography & Chest radiograph & 215 & 214 & 209 & 209 & 97.2 \\
\rowcolor{tabshade}
Radiography & Extremity radiograph, trauma & 120 & 118 & 109 & 115 & 90.8 \\
Radiography & Extremity radiograph, postoperative & 93 & 89 & 67 & 68 & 72.0 \\
\rowcolor{tabshade}
Radiography & Extremity radiograph & 29 & 28 & 11 & 23 & 37.9 \\
Radiography & Pelvis radiograph & 22 & 22 & 21 & 22 & 95.5 \\
\rowcolor{tabshade}
Radiography & Extremity radiograph, osteoarthritis & 21 & 21 & 13 & 14 & 61.9 \\
Radiography & Barium swallow & 10 & 10 & 9 & 9 & 90.0 \\
\rowcolor{tabshade}
Radiography & Cervical spine radiograph & 8 & 7 & 7 & 7 & 87.5 \\
Radiography & Lumbar spine radiograph & 8 & 7 & 6 & 6 & 75.0 \\
\rowcolor{tabshade}
Radiography & Abdominal radiograph & 6 & 6 & 5 & 5 & 83.3 \\
Radiography & Spine radiograph & 6 & 6 & 4 & 5 & 66.7 \\
\rowcolor{tabshade}
Radiography & Extremity radiograph, rheumatology & 4 & 4 & 2 & 4 & 50.0 \\
Radiography & Hemithorax radiograph & 4 & 4 & 4 & 4 & 100.0 \\
\rowcolor{tabshade}
Radiography & Pelvis radiograph, trauma & 4 & 3 & 3 & 3 & 75.0 \\
Radiography & Thoracic spine radiograph & 4 & 4 & 3 & 3 & 75.0 \\
\rowcolor{tabshade}
Radiography & Facial bones radiograph & 3 & 3 & 2 & 2 & 66.7 \\
Radiography & Fluoroscopy & 3 & 1 & 1 & 1 & 33.3 \\
\rowcolor{tabshade}
Radiography & Skull radiograph & 3 & 2 & 1 & 1 & 33.3 \\
Radiography & Arthrography & 2 & 1 & 1 & 1 & 50.0 \\
\rowcolor{tabshade}
Radiography & Mammography & 1 & 0 & 0 & 0 & 0.0 \\
Radiography & Panoramic radiograph & 1 & 0 & 0 & 0 & 0.0 \\
\rowcolor{tabshade}
Radiography & Paranasal sinuses radiograph & 1 & 1 & 0 & 1 & 0.0 \\
CT & CT extremities & 18 & 18 & 18 & 18 & 100.0 \\
\rowcolor{tabshade}
CT & CT head & 14 & 14 & 13 & 14 & 92.9 \\
CT & CT abdomen, mass & 10 & 10 & 7 & 8 & 70.0 \\
\rowcolor{tabshade}
CT & CT chest, mass & 9 & 9 & 9 & 9 & 100.0 \\
CT & CT abdomen, acute & 7 & 7 & 6 & 7 & 85.7 \\
\rowcolor{tabshade}
CT & CT abdomen, urolithiasis & 6 & 6 & 6 & 6 & 100.0 \\
CT & CT chest, lung parenchyma & 6 & 6 & 6 & 6 & 100.0 \\
\rowcolor{tabshade}
CT & CT neck, mass & 6 & 6 & 6 & 6 & 100.0 \\
CT & CT angiography whole aorta & 5 & 5 & 5 & 5 & 100.0 \\
\rowcolor{tabshade}
CT & CT chest, pulmonary embolism & 5 & 5 & 3 & 5 & 60.0 \\
CT & CT facial bones & 5 & 5 & 5 & 5 & 100.0 \\
\rowcolor{tabshade}
CT & CT head, postoperative & 3 & 3 & 2 & 3 & 66.7 \\
CT & CT bony pelvis & 2 & 2 & 1 & 1 & 50.0 \\
\rowcolor{tabshade}
CT & CT neck, inflammation & 2 & 2 & 1 & 2 & 50.0 \\
CT & CT paranasal sinuses & 2 & 2 & 1 & 2 & 50.0 \\
\rowcolor{tabshade}
CT & CT abdomen and pelvis & 1 & 1 & 1 & 1 & 100.0 \\
CT & CT abdomen, trauma & 1 & 1 & 1 & 1 & 100.0 \\
\rowcolor{tabshade}
CT & CT angiography head and neck & 1 & 1 & 0 & 0 & 0.0 \\
CT & CT angiography leg & 1 & 1 & 1 & 1 & 100.0 \\
\rowcolor{tabshade}
CT & CT cervical spine & 1 & 0 & 0 & 0 & 0.0 \\
CT & CT chest & 1 & 1 & 0 & 1 & 0.0 \\
\rowcolor{tabshade}
CT & CT chest, trauma & 1 & 1 & 1 & 1 & 100.0 \\
CT & CT coronary angiography & 1 & 1 & 1 & 1 & 100.0 \\
\rowcolor{tabshade}
CT & CT lumbar spine & 1 & 1 & 1 & 1 & 100.0 \\
CT & CT temporal bone & 1 & 1 & 1 & 1 & 100.0 \\
\rowcolor{tabshade}
MRI & MRI head & 5 & 4 & 3 & 4 & 60.0 \\
MRI & MRI knee & 5 & 4 & 4 & 4 & 80.0 \\
\rowcolor{tabshade}
MRI & MRI liver & 5 & 5 & 5 & 5 & 100.0 \\
MRI & MRI uterus & 5 & 2 & 2 & 2 & 40.0 \\
\rowcolor{tabshade}
MRI & MRI breast & 4 & 4 & 4 & 4 & 100.0 \\
MRI & MRI head, mass & 4 & 4 & 3 & 4 & 75.0 \\
\rowcolor{tabshade}
MRI & MRI hip & 4 & 2 & 1 & 1 & 25.0 \\
MRI & MRI shoulder & 3 & 2 & 2 & 2 & 66.7 \\
\rowcolor{tabshade}
MRI & MR angiography pelvis and leg & 2 & 2 & 2 & 2 & 100.0 \\
MRI & MRI bony pelvis & 2 & 2 & 2 & 2 & 100.0 \\
\rowcolor{tabshade}
MRI & MRI hand & 2 & 2 & 2 & 2 & 100.0 \\
MRI & MRI lumbar spine & 2 & 2 & 2 & 2 & 100.0 \\
\rowcolor{tabshade}
MRI & MR angiography neck & 1 & 1 & 1 & 1 & 100.0 \\
MRI & MRI ankle & 1 & 1 & 1 & 1 & 100.0 \\
\rowcolor{tabshade}
MRI & MRI bowel & 1 & 1 & 1 & 1 & 100.0 \\
MRI & MRI chest, mass & 1 & 1 & 1 & 1 & 100.0 \\
\rowcolor{tabshade}
MRI & MRI elbow & 1 & 1 & 1 & 1 & 100.0 \\
MRI & MRI head, postoperative & 1 & 1 & 1 & 1 & 100.0 \\
\rowcolor{tabshade}
MRI & MRI neck, mass & 1 & 1 & 1 & 1 & 100.0 \\
MRI & MRI prostate & 1 & 1 & 1 & 1 & 100.0 \\
\rowcolor{tabshade}
MRI & MRI wrist & 1 & 1 & 1 & 1 & 100.0 \\
Ultrasound & US breast & 13 & 4 & 4 & 4 & 30.8 \\
\rowcolor{tabshade}
Ultrasound & US abdomen & 10 & 10 & 10 & 10 & 100.0 \\
Ultrasound & US soft tissue mass & 2 & 2 & 1 & 1 & 50.0 \\
\rowcolor{tabshade}
Ultrasound & US urogenital tract & 1 & 1 & 1 & 1 & 100.0 \\
Interventional radiology & Image-guided interventions & 10 & 8 & 5 & 5 & 50.0 \\
\end{longtable}
\endgroup
\par\vspace{3pt}{\footnotesize\RaggedRight Single-region reports of the selection test set grouped by the labeled template (73 templates, 766 reports). Structured, at least one template assigned. Optimal, selected template set identical to the label. Appropriate, every selected template shares its base template with the label (includes optimal).\par}

\clearpage
\begingroup
\setlength{\tabcolsep}{2.8pt}
\renewcommand{\arraystretch}{1.20}
\fontsize{8.3}{10.4}\selectfont
\begin{longtable}{@{}>{\RaggedRight\arraybackslash}p{22.11mm}>{\RaggedRight\arraybackslash}p{61.23mm}>{\centering\arraybackslash}p{73.99mm}>{\centering\arraybackslash}p{12.76mm}@{}}
\caption{Template confusions among single-region reports}\label{tab:S3}\\
\toprule
\textbf{Modality} & \textbf{Labeled template} & \textbf{Selected template(s)} & \textbf{Reports} \\
\midrule
\endfirsthead
\multicolumn{4}{@{}l}{\sffamily\small\textbf{Table S3.} Template confusions among single-region reports (continued)}\\[4pt]
\toprule
\textbf{Modality} & \textbf{Labeled template} & \textbf{Selected template(s)} & \textbf{Reports} \\
\midrule
\endhead
\midrule\multicolumn{4}{r@{}}{\footnotesize Continued on next page}\\
\endfoot
\bottomrule
\endlastfoot
Radiography & Extremity radiograph, postoperative & Pelvis radiograph & 10 \\
\rowcolor{tabshade}
Radiography & Extremity radiograph, postoperative & Pelvis radiograph, trauma & 6 \\
Radiography & Extremity radiograph, osteoarthritis & Pelvis radiograph & 5 \\
\rowcolor{tabshade}
Radiography & Chest radiograph & US lung & 4 \\
Radiography & Extremity radiograph & Pelvis radiograph & 2 \\
\rowcolor{tabshade}
Radiography & Extremity radiograph, osteoarthritis & MRI knee & 2 \\
Radiography & Extremity radiograph, postoperative & CT extremities & 2 \\
\rowcolor{tabshade}
Radiography & Extremity radiograph, postoperative & MRI shoulder & 2 \\
Radiography & Abdominal radiograph & CT abdomen, acute & 1 \\
\rowcolor{tabshade}
Radiography & Barium swallow & Abdominal radiograph & 1 \\
Radiography & Chest radiograph & CT chest & 1 \\
\rowcolor{tabshade}
Radiography & Extremity radiograph & Pelvis radiograph, trauma & 1 \\
Radiography & Extremity radiograph & MRI shoulder & 1 \\
\rowcolor{tabshade}
Radiography & Extremity radiograph & CT extremities & 1 \\
Radiography & Extremity radiograph, postoperative & CT bony pelvis & 1 \\
\rowcolor{tabshade}
Radiography & Extremity radiograph, trauma & Pelvis radiograph, trauma & 1 \\
Radiography & Extremity radiograph, trauma & Pelvis radiograph & 1 \\
\rowcolor{tabshade}
Radiography & Extremity radiograph, trauma & CT extremities & 1 \\
Radiography & Facial bones radiograph & CT facial bones & 1 \\
\rowcolor{tabshade}
Radiography & Lumbar spine radiograph & Pelvis radiograph, trauma & 1 \\
Radiography & Skull radiograph & CT temporal bone & 1 \\
\rowcolor{tabshade}
Radiography & Spine radiograph & CT spine & 1 \\
Radiography & Thoracic spine radiograph & CT thoracic spine & 1 \\
\rowcolor{tabshade}
CT & CT abdomen, mass & CT angiography abdomen & 2 \\
CT & CT angiography head and neck & CT angiography head and neck + CT angiography & 1 \\
\rowcolor{tabshade}
CT & CT bony pelvis & CT abdomen, trauma & 1 \\
MRI & MRI hip & MRI bony pelvis & 1 \\
\rowcolor{tabshade}
Ultrasound & US soft tissue mass & US abdomen & 1 \\
Interventional radiology & Image-guided interventions & CT abdomen and pelvis & 1 \\
\rowcolor{tabshade}
Interventional radiology & Image-guided interventions & CT chest, mass & 1 \\
Interventional radiology & Image-guided interventions & CT chest & 1 \\
\end{longtable}
\endgroup
\par\vspace{3pt}{\footnotesize\RaggedRight Single-region reports that received a template outside the appropriate family (57 reports). Several selected templates are joined with a plus sign.\par}

\clearpage
\begin{table}[!htbp]
\centering
\caption{Ablation, labeled modality supplied to the pipeline}\label{tab:S4}
\begingroup
\setlength{\tabcolsep}{2.8pt}
\renewcommand{\arraystretch}{1.20}
\fontsize{8.5}{10.6}\selectfont
\begin{tabular}{@{}>{\RaggedRight\arraybackslash}p{25.96mm}>{\RaggedRight\arraybackslash}p{10.54mm}>{\centering\arraybackslash}p{22.71mm}>{\centering\arraybackslash}p{22.71mm}>{\centering\arraybackslash}p{12.17mm}>{\centering\arraybackslash}p{12.17mm}>{\centering\arraybackslash}p{44.61mm}>{\centering\arraybackslash}p{11.36mm}@{}}
\toprule
\textbf{Reports} & \textbf{n} & \textbf{Optimal, pipeline} & \textbf{Optimal, modality supplied} & \textbf{Gains} & \textbf{Losses} & \textbf{Difference, pp (95\% CI)} & \textbf{P} \\
\midrule
All & 914 & 678 & 697 & 51 & 32 & +2.1 (0.1, 4.0) & .05 \\
\rowcolor{tabshade}
Single-region & 766 & 634 & 654 & 41 & 21 & +2.6 (0.6, 4.6) & .02 \\
Multi-region & 148 & 44 & 43 & 10 & 11 & -0.7 (-6.7, 5.4) & > .99 \\
\bottomrule
\end{tabular}
\endgroup
\par\vspace{3pt}{\footnotesize\RaggedRight Optimal template selection with the labeled modality supplied to the pipeline compared with the unassisted pipeline on the same 914 reports. Gains and losses are discordant pairs, P from a two-sided exact McNemar test. pp, percentage points. The ablation is available as aggregate counts only, in which the 2 reports with the label spelling variant (Methods) are counted as errors in both arms, so the unassisted count is 678 instead of the 680 of \hyperref[tab:2]{Table~2}.\par}
\end{table}

\begin{table}[!htbp]
\centering
\caption{Ablation, reasoning effort}\label{tab:S5}
\begingroup
\setlength{\tabcolsep}{2.8pt}
\renewcommand{\arraystretch}{1.20}
\fontsize{8.5}{10.6}\selectfont
\begin{tabular}{@{}>{\RaggedRight\arraybackslash}p{22.71mm}>{\RaggedRight\arraybackslash}p{8.92mm}>{\centering\arraybackslash}p{22.71mm}>{\centering\arraybackslash}p{22.71mm}>{\centering\arraybackslash}p{11.36mm}>{\centering\arraybackslash}p{11.36mm}>{\centering\arraybackslash}p{47.86mm}>{\centering\arraybackslash}p{14.60mm}@{}}
\toprule
\textbf{Reasoning effort} & \textbf{n} & \textbf{Optimal, low effort} & \textbf{Optimal, at setting} & \textbf{Gains} & \textbf{Losses} & \textbf{Difference, pp (95\% CI)} & \textbf{P} \\
\midrule
Medium & 111 & 45 & 47 & 8 & 6 & +1.8 (-4.8, 8.4) & .79 \\
\rowcolor{tabshade}
High & 111 & 45 & 41 & 7 & 11 & -3.6 (-11.1, 3.9) & .48 \\
\bottomrule
\end{tabular}
\endgroup
\par\vspace{3pt}{\footnotesize\RaggedRight Optimal template selection at medium and high reasoning effort compared with the low effort of the pipeline on the 111 selection test reports that the pipeline routed to multi-label classification at low effort (113 reports, 2 of which combined two modalities and were not scored). Gains and losses are discordant pairs, P from a two-sided exact McNemar test. pp, percentage points.\par}
\end{table}

\FloatBarrier\clearpage
\begingroup
\setlength{\tabcolsep}{2.8pt}
\renewcommand{\arraystretch}{1.20}
\fontsize{8.3}{10.4}\selectfont
\begin{longtable}{@{}>{\RaggedRight\arraybackslash}p{23.81mm}>{\RaggedRight\arraybackslash}p{88.45mm}>{\centering\arraybackslash}p{28.92mm}>{\centering\arraybackslash}p{28.92mm}@{}}
\caption{Templates in use in the structured archive}\label{tab:S6}\\
\toprule
\textbf{Modality} & \textbf{Template} & \textbf{Structured reports} & \textbf{Share within modality, \%} \\
\midrule
\endfirsthead
\multicolumn{4}{@{}l}{\sffamily\small\textbf{Table S6.} Templates in use in the structured archive (continued)}\\[4pt]
\toprule
\textbf{Modality} & \textbf{Template} & \textbf{Structured reports} & \textbf{Share within modality, \%} \\
\midrule
\endhead
\midrule\multicolumn{4}{r@{}}{\footnotesize Continued on next page}\\
\endfoot
\bottomrule
\endlastfoot
Radiography & Chest radiograph & 493,844 & 36.5 \\
\rowcolor{tabshade}
Radiography & Extremity radiograph, trauma & 227,864 & 16.8 \\
Radiography & Extremity radiograph, postoperative & 183,504 & 13.5 \\
\rowcolor{tabshade}
Radiography & Pelvis radiograph & 83,596 & 6.2 \\
Radiography & Extremity radiograph, osteoarthritis & 56,869 & 4.2 \\
\rowcolor{tabshade}
Radiography & Pelvis radiograph, trauma & 53,686 & 4.0 \\
Radiography & Mammography & 50,403 & 3.7 \\
\rowcolor{tabshade}
Radiography & Extremity radiograph & 46,023 & 3.4 \\
Radiography & Lumbar spine radiograph & 31,210 & 2.3 \\
\rowcolor{tabshade}
Radiography & Abdominal radiograph & 22,813 & 1.7 \\
Radiography & Barium swallow & 21,750 & 1.6 \\
\rowcolor{tabshade}
Radiography & Cervical spine radiograph & 18,475 & 1.4 \\
Radiography & Hemithorax radiograph & 12,793 & 0.9 \\
\rowcolor{tabshade}
Radiography & Thoracic spine radiograph & 9,859 & 0.7 \\
Radiography & Spine radiograph & 7,062 & 0.5 \\
\rowcolor{tabshade}
Radiography & Facial bones radiograph & 6,458 & 0.5 \\
Radiography & Extremity radiograph, rheumatology & 5,686 & 0.4 \\
\rowcolor{tabshade}
Radiography & Skull radiograph & 5,221 & 0.4 \\
Radiography & Paranasal sinuses radiograph & 4,873 & 0.4 \\
\rowcolor{tabshade}
Radiography & Panoramic radiograph & 4,427 & 0.3 \\
Radiography & Contrast enema & 2,859 & 0.2 \\
\rowcolor{tabshade}
Radiography & Fluoroscopy & 2,852 & 0.2 \\
Radiography & Pelvis radiograph, rheumatology & 915 & 0.1 \\
\rowcolor{tabshade}
Radiography & Arthrography & 766 & 0.1 \\
Radiography & Urography & 683 & 0.1 \\
\rowcolor{tabshade}
CT & CT head & 106,061 & 13.6 \\
CT & CT chest, mass & 77,731 & 10.0 \\
\rowcolor{tabshade}
CT & CT abdomen, mass & 74,854 & 9.6 \\
CT & CT abdomen and pelvis & 65,059 & 8.3 \\
\rowcolor{tabshade}
CT & CT chest & 61,634 & 7.9 \\
CT & CT extremities & 57,437 & 7.4 \\
\rowcolor{tabshade}
CT & CT facial bones & 37,998 & 4.9 \\
CT & CT angiography head and neck & 28,238 & 3.6 \\
\rowcolor{tabshade}
CT & CT chest, pulmonary embolism & 27,541 & 3.5 \\
CT & CT abdomen, acute & 24,494 & 3.1 \\
\rowcolor{tabshade}
CT & CT neck & 21,665 & 2.8 \\
CT & CT angiography abdomen & 21,653 & 2.8 \\
\rowcolor{tabshade}
CT & CT chest, lung parenchyma & 21,298 & 2.7 \\
CT & CT neck, mass & 19,345 & 2.5 \\
\rowcolor{tabshade}
CT & CT angiography whole aorta & 14,400 & 1.8 \\
CT & CT abdomen, trauma & 11,292 & 1.4 \\
\rowcolor{tabshade}
CT & CT lumbar spine & 10,518 & 1.3 \\
CT & CT neck, trauma & 10,435 & 1.3 \\
\rowcolor{tabshade}
CT & CT abdomen, urolithiasis & 9,670 & 1.2 \\
CT & CT chest, trauma & 9,068 & 1.2 \\
\rowcolor{tabshade}
CT & CT cervical spine & 7,675 & 1.0 \\
CT & CT-guided biopsies, punctures and drainages & 6,870 & 0.9 \\
\rowcolor{tabshade}
CT & CT head, postoperative & 6,251 & 0.8 \\
CT & CT bony pelvis & 5,441 & 0.7 \\
\rowcolor{tabshade}
CT & CT angiography leg & 5,120 & 0.7 \\
CT & CT paranasal sinuses & 4,791 & 0.6 \\
\rowcolor{tabshade}
CT & CT neck, inflammation & 4,646 & 0.6 \\
CT & CT spine & 4,597 & 0.6 \\
\rowcolor{tabshade}
CT & CT temporal bone & 3,878 & 0.5 \\
CT & CT musculoskeletal & 3,822 & 0.5 \\
\rowcolor{tabshade}
CT & CT angiography chest & 3,121 & 0.4 \\
CT & CT thoracic spine & 3,074 & 0.4 \\
\rowcolor{tabshade}
CT & CT coronary angiography & 2,730 & 0.3 \\
CT & CT perfusion head & 2,471 & 0.3 \\
\rowcolor{tabshade}
CT & CT angiography & 1,752 & 0.2 \\
CT & CT myeloma & 1,509 & 0.2 \\
\rowcolor{tabshade}
CT & CT heart & 1,299 & 0.2 \\
CT & CT jaw & 362 & <0.1 \\
\rowcolor{tabshade}
CT & CT venography & 311 & <0.1 \\
CT & CT-guided tumor ablation & 233 & <0.1 \\
\rowcolor{tabshade}
CT & CT angiography arm & 209 & <0.1 \\
CT & Interventional pain therapy & 113 & <0.1 \\
\rowcolor{tabshade}
CT & CT polytrauma & 86 & <0.1 \\
CT & CT heart before pulmonary vein isolation & 75 & <0.1 \\
\rowcolor{tabshade}
CT & CT-guided interventional oncology & 30 & <0.1 \\
MRI & MRI breast & 30,277 & 17.9 \\
\rowcolor{tabshade}
MRI & MRI knee & 21,636 & 12.8 \\
MRI & MRI neck, mass & 11,680 & 6.9 \\
\rowcolor{tabshade}
MRI & MRI shoulder & 11,485 & 6.8 \\
MRI & MRI liver & 10,202 & 6.0 \\
\rowcolor{tabshade}
MRI & MRI head, mass & 8,889 & 5.3 \\
MRI & MRI hip & 7,956 & 4.7 \\
\rowcolor{tabshade}
MRI & MRI bony pelvis & 6,710 & 4.0 \\
MRI & MRI lumbar spine & 5,926 & 3.5 \\
\rowcolor{tabshade}
MRI & MRI ankle & 5,510 & 3.3 \\
MRI & MRI head & 4,572 & 2.7 \\
\rowcolor{tabshade}
MRI & MRI pancreas & 3,744 & 2.2 \\
MRI & MRI spine & 3,093 & 1.8 \\
\rowcolor{tabshade}
MRI & MR angiography neck & 3,084 & 1.8 \\
MRI & MRI foot & 2,913 & 1.7 \\
\rowcolor{tabshade}
MRI & MRI heart & 2,912 & 1.7 \\
MRI & MRI abdomen & 2,423 & 1.4 \\
\rowcolor{tabshade}
MRI & MRI head, acute & 2,229 & 1.3 \\
MRI & MRI head, postoperative & 2,194 & 1.3 \\
\rowcolor{tabshade}
MRI & MRI elbow & 2,194 & 1.3 \\
MRI & MRI prostate & 2,131 & 1.3 \\
\rowcolor{tabshade}
MRI & MRI hand & 2,053 & 1.2 \\
MRI & MRI wrist & 1,865 & 1.1 \\
\rowcolor{tabshade}
MRI & MRI kidney & 1,714 & 1.0 \\
MRI & MRI rectum & 1,698 & 1.0 \\
\rowcolor{tabshade}
MRI & MR cholangiopancreatography & 1,562 & 0.9 \\
MRI & MRI neck & 1,353 & 0.8 \\
\rowcolor{tabshade}
MRI & MRI cervical spine & 1,221 & 0.7 \\
MRI & MRI uterus & 1,205 & 0.7 \\
\rowcolor{tabshade}
MRI & MRI bowel & 1,103 & 0.7 \\
MRI & MRI chest, mass & 892 & 0.5 \\
\rowcolor{tabshade}
MRI & MRI thoracic spine & 750 & 0.4 \\
MRI & MRI sacroiliac joints & 613 & 0.4 \\
\rowcolor{tabshade}
MRI & MRI chest & 475 & 0.3 \\
MRI & MRI neck, trauma & 298 & 0.2 \\
\rowcolor{tabshade}
MRI & MR angiography chest & 234 & 0.1 \\
MRI & MRI ovaries & 207 & 0.1 \\
\rowcolor{tabshade}
MRI & MRI head, epilepsy & 94 & 0.1 \\
Ultrasound & US breast & 43,089 & 47.2 \\
\rowcolor{tabshade}
Ultrasound & US abdomen & 17,005 & 18.6 \\
Ultrasound & US lung & 13,465 & 14.7 \\
\rowcolor{tabshade}
Ultrasound & US joints & 7,959 & 8.7 \\
Ultrasound & US musculoskeletal (without joints) & 3,540 & 3.9 \\
\rowcolor{tabshade}
Ultrasound & US soft tissue mass & 2,802 & 3.1 \\
Ultrasound & US cervical lymph nodes & 859 & 0.9 \\
\rowcolor{tabshade}
Ultrasound & US interventional planning and follow-up & 786 & 0.9 \\
Ultrasound & US transplant follow-up & 707 & 0.8 \\
\rowcolor{tabshade}
Ultrasound & US urogenital tract & 298 & 0.3 \\
Ultrasound & US trauma (FAST) & 218 & 0.2 \\
\rowcolor{tabshade}
Ultrasound & US thyroid & 166 & 0.2 \\
Ultrasound & US leg veins (thrombosis) & 145 & 0.2 \\
\rowcolor{tabshade}
Ultrasound & US shunt and transplant vessels & 65 & 0.1 \\
Ultrasound & US leg arteries (peripheral artery disease, aneurysm, inflammation) & 55 & 0.1 \\
\rowcolor{tabshade}
Ultrasound & US salivary glands & 28 & <0.1 \\
Ultrasound & US hernia & 26 & <0.1 \\
\rowcolor{tabshade}
Ultrasound & US gynecological (transvaginal or transabdominal) & 23 & <0.1 \\
Ultrasound & US arm veins (thrombosis) & 23 & <0.1 \\
\rowcolor{tabshade}
Ultrasound & US nerve compression & 13 & <0.1 \\
Ultrasound & Transthoracic echocardiography & 13 & <0.1 \\
\rowcolor{tabshade}
Ultrasound & US hands rheumatology & 12 & <0.1 \\
Ultrasound & US contrast-enhanced (CEUS) & 11 & <0.1 \\
\rowcolor{tabshade}
Ultrasound & US extracranial cerebral arteries & 8 & <0.1 \\
Ultrasound & US arm arteries (stenosis, aneurysm, inflammation) & 7 & <0.1 \\
\rowcolor{tabshade}
Ultrasound & US abdominal aorta & 1 & <0.1 \\
Ultrasound & US pediatric abdomen & 1 & <0.1 \\
\rowcolor{tabshade}
Interventional radiology & Image-guided interventions & 4,755 & 82.4 \\
Interventional radiology & Interventional oncology & 462 & 8.0 \\
\rowcolor{tabshade}
Interventional radiology & Venous interventions & 286 & 5.0 \\
Interventional radiology & Tumor ablation & 163 & 2.8 \\
\rowcolor{tabshade}
Interventional radiology & Interventional pain therapy & 88 & 1.5 \\
Interventional radiology & Arterial interventions & 14 & 0.2 \\
\rowcolor{tabshade}
Interventional radiology & Vascular interventions & 5 & 0.1 \\
Interventional radiology & Tumor embolization & 1 & <0.1 \\
\end{longtable}
\endgroup
\par\vspace{3pt}{\footnotesize\RaggedRight All 142 templates that received at least one structured report in the TUM archive, with the number of structured reports and the share within the modality. The interventional pain therapy template is listed under CT and interventional radiology.\par}

\FloatBarrier
\clearpage\subsection{Supplemental figures}
\begin{figure}[!htbp]
\centering
\includegraphics[width=\linewidth]{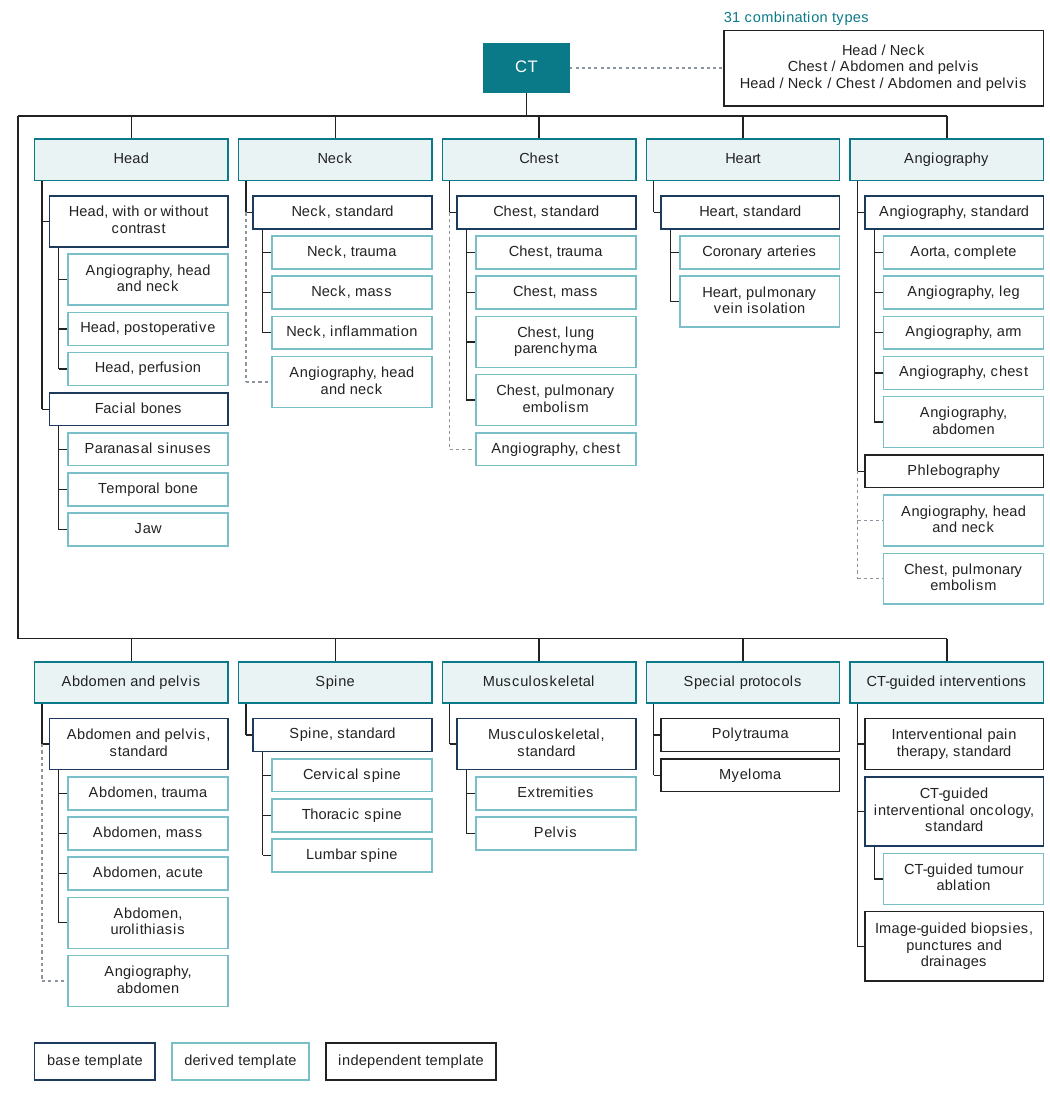}
\caption{Template index of the CT branch, the largest of the five modality branches (template counts per modality in \hyperref[tab:1]{Table~1}, templates in archive use in \hyperref[tab:S6]{Table~S6}). Ten examination types with 45 templates, five of which are listed under two or three examination types and appear under each. Base templates (navy outline) carry derived templates (light teal outline) that add or replace sections at run time, independent templates (black outline) have no base. Templates derived from a base template of another examination type are attached with dashed connectors. Three of the 31 combination types are shown as examples.}\label{fig:S1}
\end{figure}

\clearpage
\begin{figure}[!htbp]
\centering
\includegraphics[width=\linewidth]{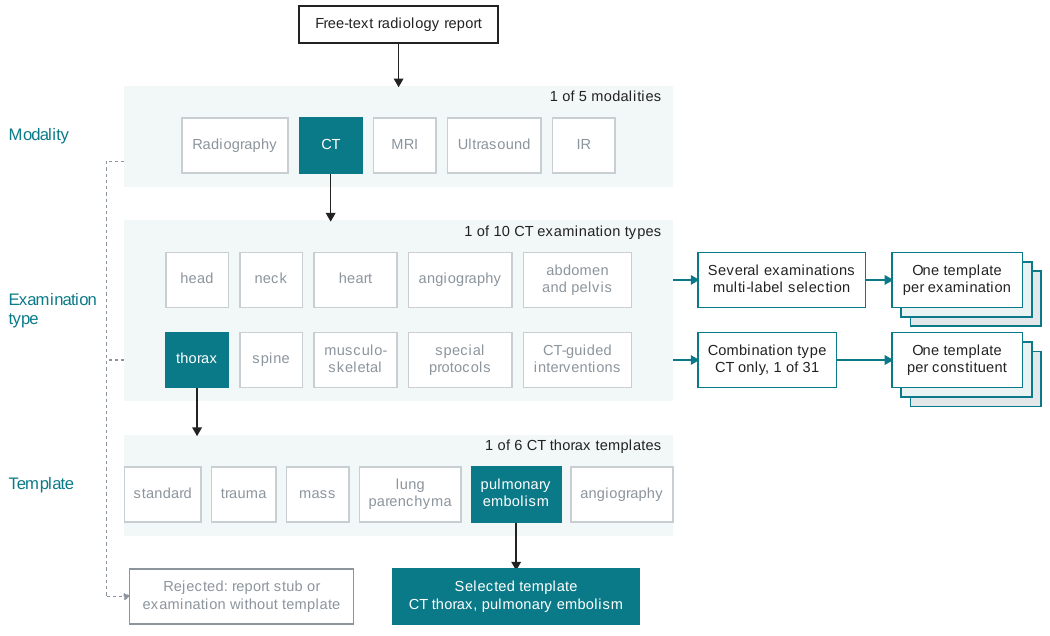}
\caption{Hierarchical template selection as a tree. The modality step chooses one of 5 modalities, which restricts the examination-type step to the examination types of that modality (10 for CT), which restricts the template step to the templates of that examination type (6 for CT chest). Teal branches show the multi-label pathways for reports describing several examinations and for CT combination types. Dashed grey exits show the two reserved rejection responses.}\label{fig:S2}
\end{figure}

\clearpage
\begin{figure}[!htbp]
\centering
\includegraphics[width=\linewidth]{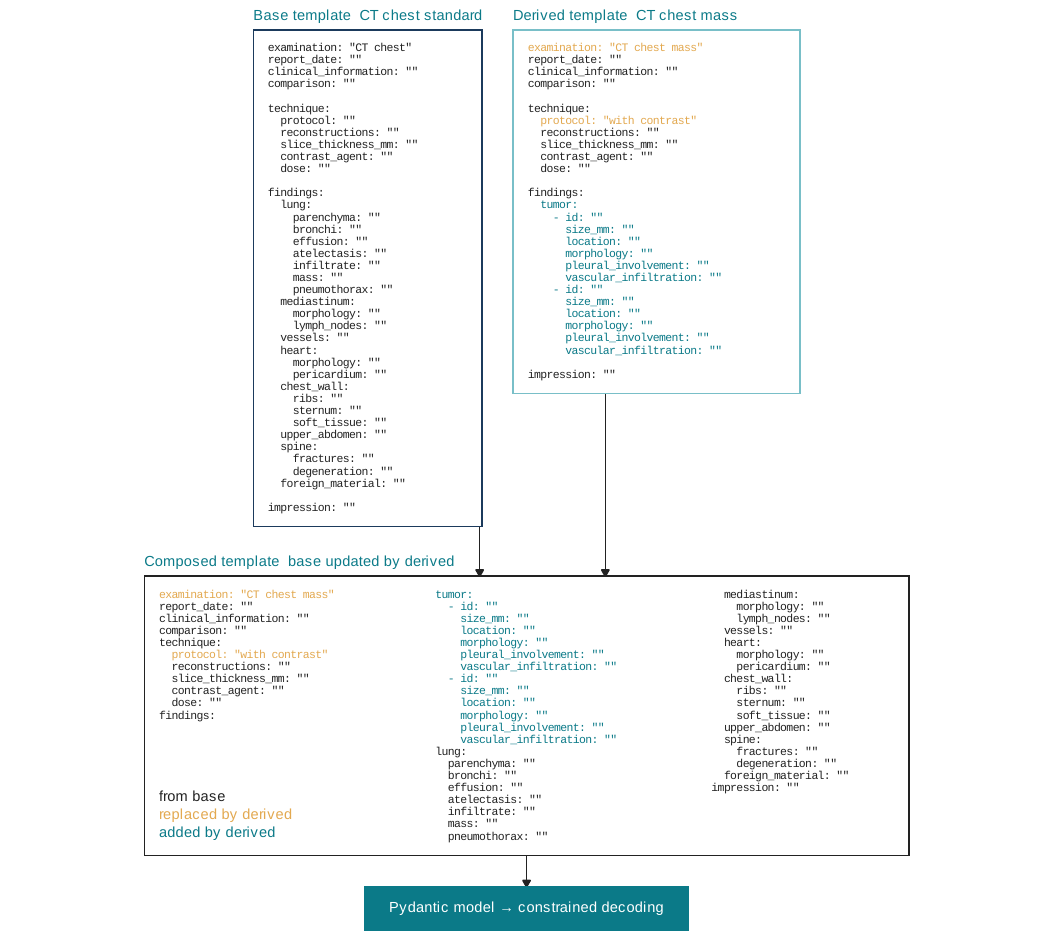}
\caption{Template composition at run time with the YAML files of the base template CT chest standard and the derived template CT chest mass. The production templates are German, keys and values are translated one to one for display and the placeholder hints of the production files are shown as empty strings. The pipeline updates the base template recursively with the derived template, a leaf value in the derived template replaces the base value (sand), sections present only in the derived template are added (teal), all other sections remain from the base (black). The composed template becomes the Pydantic model for constrained decoding.}\label{fig:S3}
\end{figure}

\clearpage
\section{Supplement 1. CLAIM 2024 checklist}\label{supp:claim}
Checklist for Artificial Intelligence in Medical Imaging (CLAIM), 2024 update, 44 items. Locations refer to section headings in this preprint. Items marked No or NA carry an explanation. Thirty-five items are reported, one is not reported (item 32, demographic characteristics unavailable at report level), and eight are not applicable.

\setcounter{table}{6}
\begingroup
\setlength{\tabcolsep}{2.8pt}
\renewcommand{\arraystretch}{1.20}
\fontsize{8.2}{10.299999999999999}\selectfont
\begin{longtable}{@{}>{\RaggedRight\arraybackslash}p{7.57mm}>{\RaggedRight\arraybackslash}p{28.58mm}>{\RaggedRight\arraybackslash}p{12.61mm}>{\RaggedRight\arraybackslash}p{39.51mm}>{\RaggedRight\arraybackslash}p{79.86mm}@{}}
\caption{CLAIM 2024 checklist}\label{tab:S7}\\
\toprule
\textbf{Item} & \textbf{Topic} & \textbf{Reported} & \textbf{Location in manuscript} & \textbf{Notes} \\
\midrule
\endfirsthead
\multicolumn{5}{@{}l}{\sffamily\small\textbf{Table S7.} CLAIM 2024 checklist (continued)}\\[4pt]
\toprule
\textbf{Item} & \textbf{Topic} & \textbf{Reported} & \textbf{Location in manuscript} & \textbf{Notes} \\
\midrule
\endhead
\midrule\multicolumn{5}{r@{}}{\footnotesize Continued on next page}\\
\endfoot
\bottomrule
\endlastfoot
1 & Identification as AI study & Yes & Title, Abstract & Title and abstract name the open-weight large language model pipeline. \\
\rowcolor{tabshade}
2 & Structured summary & Yes & Abstract & Retrospective design, data source (two centers), 914 and 920 test reports, five modalities, archive of 2,186,982 reports, metrics with 95\% CIs. \\
3 & Scientific and clinical background & Yes & Introduction, paragraphs 1 to 3 & \textemdash \\
\rowcolor{tabshade}
4 & Study objectives and hypotheses & Yes & Introduction, last paragraph & Objective stated, no hypothesis (retrospective study), not registered. \\
5 & Prospective or retrospective & Yes & Methods, Study Design & Retrospective. \\
\rowcolor{tabshade}
6 & Study goal and AI task & Yes & Introduction, last paragraph; Methods, Pipeline & Hierarchical classification of the template and zero-shot structuring of report text, intended as autonomous post hoc structuring of archived reports without human oversight. \\
7 & Data source & Yes & Methods, Study Design and Test Datasets & Charité archive for template development and prompt engineering, TUM archive for the two test sets and the complete archive. The reports had not been used in prior studies. \\
\rowcolor{tabshade}
8 & Eligibility criteria & Yes & Methods, Test Datasets and Archive Processing; \hyperref[fig:1]{Figure~1}; \hyperref[tab:4]{Table~4} note & Counts per exclusion criterion in \hyperref[fig:1]{Figure~1} (test sets) and \hyperref[tab:4]{Table~4} note (archive). \\
9 & Data partitioning & Yes & Methods, Study Design and Test Datasets & No model training. Development (templates, prompts) at Charité, all testing at TUM, partition at report level. \\
\rowcolor{tabshade}
10 & Data preprocessing & Yes & Methods, Pipeline and Archive Processing & Report text is used as written, entries without identifier or with a foreign-image header are excluded. \\
11 & De-identification & Yes & Methods, Study Design and Model; Data sharing statement & Reports were processed on hospital hardware inside the institution, no text left the institution, report texts are not shared. \\
\rowcolor{tabshade}
12 & Missing data & Yes & Methods, Test Datasets; \hyperref[fig:1]{Figure~1}; \hyperref[tab:4]{Table~4} note & Entries without a report identifier (21 in the sample, 47,807 in the archive) were removed. \\
13 & Image acquisition & NA & \textemdash & Text-only study, no images were used. \\
\rowcolor{tabshade}
14 & Reference standard definition & Yes & Methods, Test Datasets and Evaluation Metrics & Expert labels (modality, examination types, optimal templates, non-structurable) for selection, resident-corrected structured output for structuring. \\
15 & Rationale for reference standard & Yes & Methods, Test Datasets and Evaluation Metrics & Field-by-field correction of the pipeline output against the source report yields a reference in the template structure so that the template text does not contribute to the scores. \\
\rowcolor{tabshade}
16 & Annotation tools & Yes & Methods, Test Datasets & Labels and corrections were entered in software developed by the authors for this study. \\
17 & Human annotators & Yes & Methods, Test Datasets & One board-certified radiologist labeled, a second reviewed all labels, five residents (2 to 3 years of experience) rated and corrected, two board-certified radiologists adjudicated unclear ratings in consensus. Inter-rater variability was not assessed. \\
\rowcolor{tabshade}
18 & Model architecture & Yes & Methods, Model and Pipeline & gpt-oss-120B, 4-bit quantized mixture-of-experts model, 117 billion total and 5.1 billion active parameters, used with constrained decoding against 150 hierarchically organized templates. \\
19 & Pre-trained model details & Yes & Methods, Model & Open-weight pretrained model used zero-shot without fine-tuning. \\
\rowcolor{tabshade}
20 & Inputs and outputs & Yes & Methods, Pipeline; Supplemental Methods S1 and S3 & Input report text and response format, output one option per selection step and one JSON document per selected template. \\
21 & Feature extraction & NA & \textemdash & No engineered features. \\
\rowcolor{tabshade}
22 & Feature selection & NA & \textemdash & No engineered features. \\
23 & Class imbalance handling & NA & \textemdash & No model training. \\
\rowcolor{tabshade}
24 & Uncertainty quantification & NA & \textemdash & No uncertainty estimates. Semantic textual similarity between output and source is discussed as a post hoc flag for review (Discussion). \\
25 & Training approach & NA & \textemdash & No model training, zero-shot use of a pretrained model. \\
\rowcolor{tabshade}
26 & Hyperparameter tuning & Yes & Methods, Study Design and Model; \hyperref[tab:S5]{Table~S5} & Prompts were developed at Charité, reasoning effort set to low, context window 15,000 tokens, medium and high reasoning effort tested as an ablation. \\
27 & Software and libraries & Yes & Methods, Pipeline, Model and Evaluation & Python, LangChain 1.2.18, vLLM, Ollama, Llama-Embed-Nemotron-8B, SciPy. \\
\rowcolor{tabshade}
28 & Evaluation metrics & Yes & Methods, Evaluation Metrics and Statistical Analysis & Sensitivity, specificity, optimal and appropriate selection, STS, BLEU, field-level correctness, hallucination rate. \\
29 & Statistical analysis & Yes & Methods, Evaluation Metrics and Statistical Analysis & Wilson 95\% CIs, percentile bootstrap (10,000 resamples), exact McNemar tests, P < .05, Python with SciPy. \\
\rowcolor{tabshade}
30 & Reader study design & Yes & Methods, Test Datasets & Readers rated and corrected the pipeline output, no comparison of AI against readers. \\
31 & Study flow diagram & Yes & \hyperref[fig:1]{Figure~1} (test sets), \hyperref[fig:6]{Figure~6A} (archive) & \textemdash \\
\rowcolor{tabshade}
32 & Demographic characteristics & No & Methods, Study Design; \hyperref[tab:1]{Table~1} & Patient demographics were not available at report level, as stated in Methods. Modality, examination types, region count and report length per test set are given in \hyperref[tab:1]{Table~1}. \\
33 & Performance summary & Yes & Results; Tables 2 to 4 & No training or validation partitions, all results are test results. \\
\rowcolor{tabshade}
34 & Failures and errors & Yes & Results, Archive-Scale Processing; \hyperref[tab:4]{Table~4} note; \hyperref[fig:6]{Figure~6A} & 75,452 archive reports without output by cause, regeneration of unparseable responses (Supplemental Methods S3). \\
35 & Test set performance & Yes & Results; Tables 2 and 3; Figures 3 to 5 & All proportions with 95\% CIs. \\
\rowcolor{tabshade}
36 & Reader comparison & NA & \textemdash & No comparison of AI against human readers. \\
37 & Study limitations & Yes & Discussion, limitations paragraph & \textemdash \\
\rowcolor{tabshade}
38 & Clinical applicability & Yes & Discussion & Cost and feasibility of archive-scale local processing, use of the structured archive, template adaptation by other institutions. \\
39 & Safety and bias considerations & Yes & Discussion & Hallucination and omission rates, flagging of low-similarity outputs for review, language dependence. Demographic subgroups were not analyzed. \\
\rowcolor{tabshade}
40 & Registration & NA & \textemdash & Retrospective study. \\
41 & Conflicts of interest & Yes & Declarations, Conflicts of interest & Research collaborations, grant support, speaker fees, and consultant fees are disclosed. \\
\rowcolor{tabshade}
42 & AI use disclosure & Yes & Declarations, Use of artificial intelligence & \textemdash \\
43 & Code and data availability & Yes & Declarations, Data sharing statement & Pipeline, templates, index, prompts and evaluation scripts on GitHub, report texts not shareable. \\
\rowcolor{tabshade}
44 & Funding and support & Yes & Declarations, Funding & European Union: COMFORT (101079894) and TWIN-X (101287679). The funder had no role in study design, data collection, analysis, or writing. \\
\end{longtable}
\endgroup

\end{document}